\makeatletter
\let\@twosidetrue\@twosidefalse
\let\@mparswitchtrue\@mparswitchfalse
\makeatother

\documentclass[runningheads]{llncs}
\usepackage{graphicx}
\usepackage{comment}
\usepackage{amsmath,amssymb} 
\usepackage{color}

\usepackage[margin=1in]{geometry}

\usepackage{subfigure} 

\usepackage[table]{xcolor}
\usepackage{makecell}

\newcommand{\ignore}[1]{}

\usepackage{hyperref}
\hypersetup{breaklinks=true,letterpaper=true,colorlinks}

\DeclareMathOperator*{\argmax}{arg\,max}

\begin{document}

\pagestyle{headings}
\mainmatter

\title{Detecting Deep-Fake Videos from \\  Appearance and Behavior}

\titlerunning{Detecting Deep Fakes}

\author{Shruti Agarwal\inst{1}, Tarek El-Gaaly\inst{2}, Hany Farid\inst{1}, \and Ser-Nam Lim\inst{2}}
\authorrunning{S. Agarwal et al.}

\institute{Univeristy of California, Berkeley, Berkeley, CA, USA \and
Facebook Research, New York, NY, USA
}
%

\maketitle

\begin{abstract}
Synthetically-generated audios and videos -- so-called deep fakes -- continue to capture the imagination of the computer-graphics and computer-vision communities. At the same time, the democratization of access to technology that can create sophisticated manipulated video of anybody saying anything continues to be of concern because of its power to disrupt democratic elections, commit small to large-scale fraud, fuel dis-information campaigns, and create non-consensual pornography. We describe a biometric-based forensic technique for detecting face-swap deep fakes. This technique combines a static biometric based on facial recognition with a temporal, behavioral biometric based on facial expressions and head movements, where the behavioral embedding is learned using a CNN with a metric-learning objective function. We show the efficacy of this approach across several large-scale video datasets, as well as in-the-wild deep fakes.

\keywords{Digital Forensics, AI-Synthesized Content, Deep Fakes}
\end{abstract}

\section{Introduction}
\label{sec: intro}

Recent advances in computer graphics, computer vision, and machine learning have made it increasingly easier to synthesize compelling fake audio, image, and video. In the audio domain, highly realistic audio synthesis is now possible in which a neural network, with enough sample recordings, can learn to synthesize speech in your voice~\cite{oord2016}. In the static image domain, highly realistic images of people can now be synthesized using generative adversarial networks (GANs)~\cite{karras2019style1,karras2019style2}. And, in the video domain, highly realistic videos can be created of anybody saying and doing anything that its creator wants~\cite{tolosana2020deepfakes}. These so-called deep-fake videos can be highly entertaining but can also be easily weaponized. 

The creation of non-consensual pornography, for example, was the first use of deep fakes, and continues to pose a threat particularly to women, ranging from celebrities to journalists, and those that simply attract unwanted attention~\cite{chesney2019}. In response, several U.S. states have recently passed legislation trying to mitigate the harm posed by this content, and similar legislation is being considered at the U.S. federal and international levels. In addition, the democratization of access to sophisticated technology to synthesize highly realistic fake audio, image, and videos promises to add to our struggle to contend with dis- and mis-information campaigns designed to commit small- to large-scale fraud, disrupt democratic elections, and sow civil unrest.

We describe a forensic technique to authenticate face-swap deep fake videos in which a person's facial identity is replaced with another's. The most common approach to detecting these deep fakes leverages low-level pixel artifacts introduced during the synthesis process. These approaches suffer from vulnerability to simple counter-measures including trans-coding and resizing, and often struggle to generalize to new synthesis techniques (see Section~\ref{sec:related_work} for more details).

In contrast, in our approach we leverage a more fundamental flaw in deep fakes: the face-swap deep fake is simply not the person it purports to be. In particular, we combine a static biometric based on facial identity with a temporal, behavioral biometric based on facial expressions and head movements. The former leverages standard techniques from face recognition, while the latter leverages a learned behavioral embedding using a convolutional neural network (CNN) powered by a metric-learning objective function. These two biometric signals are used because we observe that the facial behaviors in a face-swap deep fake remain those of the original individual, while the facial identity is of a different individual. By matching the behavioral and facial identities against a set of authentic reference videos, inconsistencies in the matching identities can reveal face-swap deep fakes. Our experimental results against thousands of unique identities spanning five large datasets support this hypothesis.

Our behavioral model is constructed by stacking together static FAb-Net features~\cite{wiles2018} over time (four seconds). By combining many FAb-Net features, which themselves capture static head pose, facial landmarks, and facial expression, we are able to capture spatiotemporal behaviors. Unlike previous work for modeling spatiotemporal human behavior~\cite{agarwal2019} that required a specific model for each person, we will show that the metric-learning objective used by our CNN to learn this behavioral feature allows us to build a generic model that can be trained on one group of people in one dataset and generalize to previously unseen people in different datasets. We summarize our primary contributions as:
\begin{itemize}
    \item[$\bullet$] a novel spatiotemporal behavior model for capturing facial expressions and head movement that generalizes to previously unseen people;
    \item[$\bullet$] a novel combination of appearance and behavioral biometrics for detecting face-swap deep fake videos;
    \item[$\bullet$] a large-scale evaluation across five large data sets consisting of thousands of real and deep-fake videos, the results of which show that our approach is highly effective at detecting face-swap deep fakes; and
    \item[$\bullet$] an analysis of the underlying methodology and results that provides insight into the specific nature of the learned features, and the robustness of our approach across different datasets, manipulations, and qualities of deep fakes.
\end{itemize}

In the next section, we place our work in context relative to previous efforts. We then describe our technique in detail and show the efficacy of our approach across five large-scale video datasets, as well as in-the-wild deep fakes.

\section{Related Work}
\label{sec:related_work}

We begin by describing the most relevant work in both the creation and detection of deep fakes. 

\subsection{Generating Deep Fakes}

The term deep fake is often used to describe synthetically-generated images and videos (typically of people) generated from a range of different techniques including DeepFake FaceSwap~\cite{deepfakes_faceswap} and FS-GAN~\cite{nirkin2019}, Neural Textures~\cite{sies2019}, Face2Face~\cite{thies2016face2face}, and FaceSwaps~\cite{faceswap}.

The popular DeepFake FaceSwap software uses a generative adversarial network (GAN)~\cite{goodfellow14} to create so-called face-swap deep fakes in which one person's identity in a video is replaced with another person's identity. This approach has been popularized by, for example, adding the actor Nicholas Cage into movies in which he never appeared, including his highly entertaining appearance in The Sound of Music~\footnote{youtu.be/MHkZEpfUnAA}. While this technique can generate highly convincing fakes, it often requires a significant amount of training data. Faceswap-GAN~\cite{faceswap_gan}, uses an autoencoder with an adversarial loss to generate deep fakes,  similarly requiring large amounts of training data. The more recent method FS-GAN~\cite{nirkin2019}, on the other hand, creates high-quality fakes using a recurrent neural network-based reenactment with less training data. 

Neural Textures~\cite{sies2019} is a generic image synthesis framework that combines traditional graphics rendering with more modern learnable components. This framework can be used for novel-view synthesis, scene editing, and the creation of so-called lip-sync deep fakes in which a person's mouth is modified to be consistent with a new audio track. This work generalizes earlier work that was designed to create lip-sync deep fakes on a per-individual basis~\cite{suwajanakorn2017}.

Unlike these learning-based methods, some methods rely on more traditional computer-graphics approaches to create deep fakes. Face2Face~\cite{thies2016face2face}, for example, allows for the creation of so-called \textit{puppet-master} deep fakes in which one person's (the master's) facial expressions and head movements are mapped onto another person (the puppet). Similarly FaceSwap~\cite{faceswap} builds a $3$-D facial model of one person and aligns this to another person. These techniques allow for re-enacting facial expressions in real-time using standard consumer cameras. In a related puppet-master technique, the authors in~\cite{Koki2018} build a photo-realistic avatar GAN that synthesizes faces in arbitrary expressions and orientations in real-time on a mobile device. 

\subsection{Detecting Deep Fakes}

There is a significant literature in the general area of digital forensics~\cite{farid2016}. Here we focus only on techniques for detecting the types of deep-fake videos described in the previous section. 

\noindent\textbf{Low-level approaches} focus on detecting pixel-level artifacts introduced by the synthesis process. One such approach uses a CNN to detect pixel-level artifacts that arise due to the process of warping a face region onto the target~\cite{li2018warping}. In~\cite{Huh2018}, the authors train a Siamese network to find inconsistencies of camera metadata for small image patch (e.g.,~focal length, ISO, aperture size, exposure time, etc.). An image is then authenticated using this network to determine if each image patch is consistent with the same imaging pipeline. Although not necessarily focused on deep fakes, Mantra-Net~\cite{Wu2019} uses end-to-end training of a fully convolutional network to detect and localize different types of image manipulation including splicing, removal, and copy-move. In~\cite{zhou2017}, the authors detect and localize facial manipulations by using a network to  holistically classify a face as manipulated or not. A second network exploits low-level steganographic features in small patches to determine if a face region is consistent with the rest of the image. A final prediction is generated by combining these two predictions. The authors in~\cite{yu2018} and~\cite{zhang2019} showed that GAN-generated content contains distinct digital fingerprints which can be learned and used to classify images as GAN-generated or not. MesoNet~\cite{afchar2018} take a mid-level approach by building a convolutional neural network with a small number of layers that learns mesoscopic artifacts. These mid-level artifacts tend to be more resilient, particularly to video compression.

The benefit of these and similar low-level approaches is that they can automatically extract artifacts and differences between synthetic and real content. The drawback is that they can be highly sensitive to intentional or unintentional laundering including resizing or trans-coding, as well as adversarial attacks~\cite{carlini2016evaluating} and extrapolation to novel datasets. In contrast, the high-level approaches described next tend to be more resilient to these types of laundering and attacks and more likely to generalize to novel datasets. \\

\noindent\textbf{High-level approaches} focus on more semantically meaningful features. For example,~\cite{yang2019} recognized that the creation of face-swap deep fakes introduces inconsistencies in the head pose as estimated from the central, swapped portion of the face and the surrounding, original head. These inconsistencies leverage $3$-D geometry which are currently difficult for synthesis techniques to correct. Because training data sets often do not depict people with their eyes closed, it was observed that early face-swap deep fakes contained an abnormally low number of eye blinks~\cite{li2018blinking}. More recent deep fakes, however, seem to have corrected for this problem. A related technique~\cite{ciftci2019} exploits spatial and temporal physiological signals that appear not to be consistent across real videos and disrupted in face-swap deep fakes. We believe that, because current synthesis techniques are frame-based, incorporating these types of semantic and temporal dynamics is essential to staying slightly ahead of the cat-and-mouse game of synthesis and detection.

The work of~\cite{agarwal2019} is most similar to ours. In their work, the authors analyzed hours of video of specific individuals (in their case, various world leaders and presidential candidates) in order to extract distinct and predictable patterns of facial expressions and head movements. Specifically, from each $10$-second clip of an individual, the authors extracted the frame-by-frame facial expressions (parameterized as $18$ action units~\cite{ekman1976} and $3$-D head rotation about two axes). The correlation between all pairs of these $20$ features yielded a $190$-D feature vector capturing an individual's temporal mannerisms. A one-class SVM~\cite{scholkopf2001} was employed to classify each $10$-second video clip as being consistent or not with the learned mannerisms of an individual. The benefit of this approach is that it captures temporal mannerisms that current frame-based, deep-fake synthesis techniques are not (yet) able to synthesize. The other benefit is that this approach, unlike  pixel-based detection schemes, is more robust to laundering attacks and is more able to generalize to a large class of deep fakes from face-swap to lip-sync, and puppet-master. The drawback of this approach is that it can require significant effort to create models for each individual and it is almost certainly the case that the hand-crafted correlation-based features are not optimal, nor are they capturing all of the distinct properties that might distinguish a real from a fake video.

Building on this earlier work by~\cite{agarwal2019}, we employ a convolutional neural network (CNN) with a metric-learning objective function to learn a more discriminating behavioral biometric. We pair this learned biometric with a facial biometric in order to determine if a person's identity in video clips as short as four seconds is consistent with the facial and behavioral properties extracted from reference videos. This approach is specifically targeted towards face-swap deep fakes in which the face of one person has been replaced with another.


%
%
\begin{figure}[t]
    \centering
    \includegraphics[width=1\textwidth]{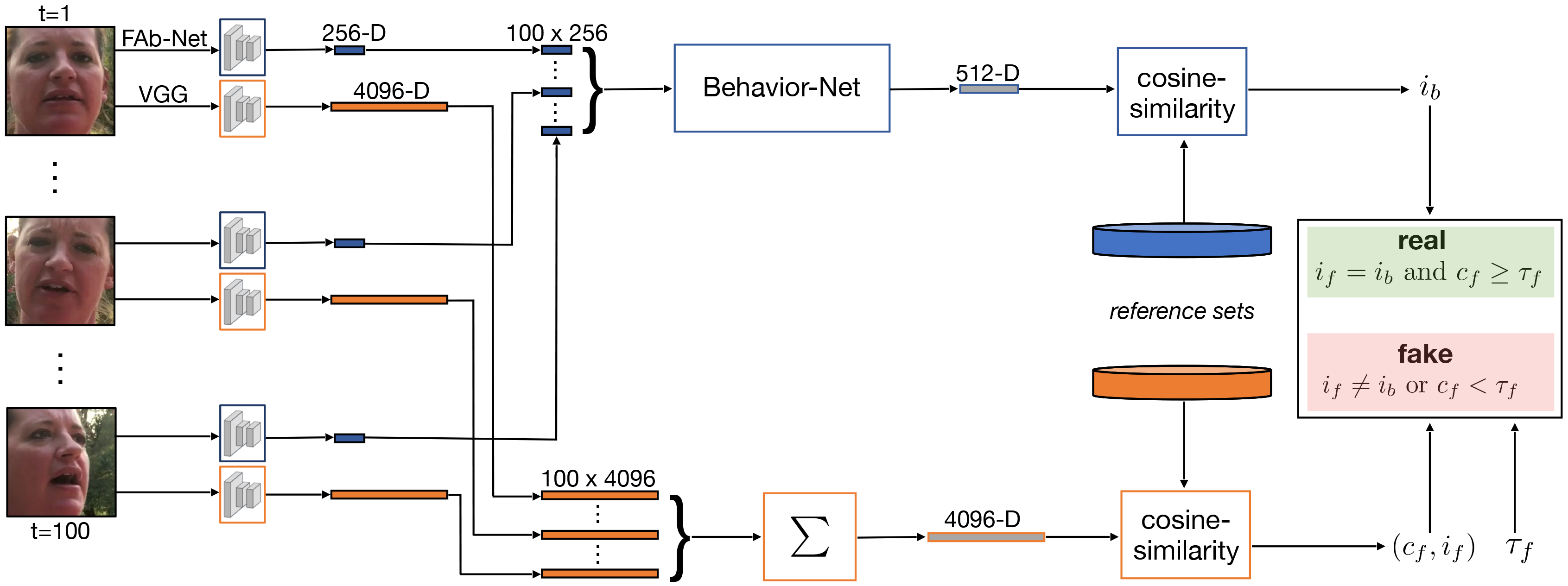}
    \caption{An overview of our authentication pipeline (see Section~\ref{sec:authentication}).}
    \label{fig:overview}
\end{figure}

\section{Biometrics}
\label{sec:biometrics}

We next describe two biometric measurements that underlie our forensic detection scheme. These include a biometric based on  temporal behavioral (facial expressions and head movements) and a biometric based on static facial features.

\subsection{Behavior}
\label{sec:behavior}

In~\cite{wiles2018}, the authors proposed a self-supervised, encoder-decoder network (Facial Attributes-Net, FAb-Net) trained to embed the movement between video frames into a common $256$-D space. The authors showed that the network, in turn, learns an embedding space that represents head pose, facial landmarks, and facial expression. We use these $256$-D FAb-Net features as building blocks to measure spatiotemporal biometric behavior. Specifically, a $t$-frame video clip of a person talking is first reduced to a feature matrix $X \in \mathbb{R}^{256 \times t}$, where each matrix column corresponds to each frame's FAb-Net feature.

FAb-Net nicely captures the frame-based facial movements and expressions but is, by design, identity-agnostic. We seek to learn a modified embedding that both captures facial movements and expressions, but also distinguishes these features across individuals. That is, starting with the static FAb-Net features, we learn a low-dimensional mapping that encodes identity-specific spatiotemporal behavior. 

Given FAb-Net feature matrices for $n$, $t$-frame video clips $X_1, \ldots, X_n$ with identity labels $y_1, \ldots, y_n$, we learn a mapping $f(\cdot): \mathbb{R}^{256 \times t} \rightarrow \mathbb{R}^d$, that projects $X_i$ to an embedding space such that the similarity $S_{ij}$ between $f(X_i)$ and $f(X_j)$ is high if $y_i=y_j$ (positive sample) and $S_{ij}$ is low if $y_i \neq y_j$ (negative sample). Because, the output $f(X_i)$ is normalized to lie on a unit sphere, a cosine similarity, between two vector-based representations, is used to compute $S_{ij}$.

To learn the mapping $f(\cdot)$, a CNN is trained with a multi-similarity metric-learning objective function~\cite{wang2019}. Following the approach in~\cite{wang2019}, the loss for a mini-batch is computed as follows. First, for every input $X_i$, hard positive and negative samples are selected. For hard negative samples (where $y_i \neq y_j$), a sample $X_j$ is selected if $S_{ij} > \min\{S_{ik} - \epsilon\}$, for all $k$ such that $y_i = y_k$, and where $\epsilon$ is a small margin. This formulation selects the most confusing negative samples whose similarity with the input is larger than the minimum similarity between the input and all positive samples. Similarly, for hard positive samples (where $y_i = y_j$), a sample $X_j$ is selected if $S_{ij} < \max\{S_{ik} + \epsilon\}$, for all $k$ such that $y_i \neq y_k$. Here, the most meaningful positive samples are selected by comparing to the negative samples most similar to the input. 

A soft weighting is then applied to rank these selected samples according to their importance for learning the desired embedding space. For a given input $X_i$, let $\mathcal{N}_i$ and $\mathcal{P}_i$ represents the selected negative and positive samples that are weighted as follows:
\begin{equation} 
    w_{ij}^{-} ~=~ \frac{e^{\beta(S_{ij}-\lambda)}}{1+\sum_{k \in \mathcal{N}_i} e^{\beta (S_{ik}-\lambda)}} \quad \mbox{and} \quad
    w_{ij}^{+} ~=~ \frac{e^{-\alpha(S_{ij}-\lambda)}}{1+\sum_{k \in \mathcal{P}_i} e^{-\alpha (S_{ik}-\lambda)}},
\end{equation}
where $\alpha$, $\beta$, and $\lambda$ are hyper-parameters. Finally, the loss $\mathcal{L}$ over a mini-batch of size $m$ is:
\begin{equation}
    \mathcal{L} ~=~ \frac{1}{m} \sum_{i=1}^{m} \left\{ \frac{1}{\alpha} \log \left[ 1 + \sum_{k \in \mathcal{P}_i} e^{-\alpha(S_{ik} - \lambda)} \right] + \frac{1}{\beta} \log \left[ 1 + \sum_{k \in \mathcal{N}_i} e^{\beta(S_{ik} - \lambda)} \right] \right\}.
\end{equation} 
By performing supervised training using the identity labels in the training data, the network is encouraged to learn an embedding space that clusters the biometric signatures by identity.

Our model is trained on the VoxCeleb2 dataset~\cite{voxceleb2}, containing over a million utterances from $5,994$ unique identities. The size of the input feature matrix is fixed to $t=100$, corresponding to a $4$-second video clip at $25$ frames/second (this clip size was selected as it was the minimum clip size of the VoxCeleb2 utterances). We used the ResNet-101 network architecture~\cite{he2016deep}, where the input layer of the network is modified to the size of our feature matrix ($256 \times 100$).  A fully-connected output layer of size $d=512$ is added on top of this network, forming our final feature vector, which is normalized to be zero-mean and unit-length before computing the loss. We name this network Behavior-Net.

The CNN training is performed for $10,000$ iterations with a mini-batch of size $256$. Following~\cite{wang2019}, in each mini-batch, $32$ identities are randomly selected, for which eight utterance videos (each of variable length) are randomly selected, from which a randomly selected $100$-frame sequence is extracted. All other optimization hyper-parameters are the same as in~\cite{wang2019}.

Even though the Behavior-Net features are trained only on the VoxCeleb2 dataset, as described below, these features will be used to classify different identities across different datasets. This generalizability is both practically useful and suggests that the underlying Behavior-Net captures intrinsic properties of people.

\subsection{Appearance}
\label{sec:appearance}

Rapid advances in deep learning and access to large datasets have led to a revolution in face recognition. We leverage one such fairly straight-forward approach, VGG~\cite{parkhi2015}, a $16$-layer CNN trained to perform face recognition on a dataset consisting of $2,622$ identities. VGG yields a distinct $4096$-D face descriptor per face, per video frame. These descriptors are averaged over the $100$ frames of the $4$-second video clip to yield a single facial descriptor.

Faces for this facial biometric and the behavioral biometric are extracted using OpenFace~\cite{baltruvsaitis2016}. Once localized and extracted from a video frame, each face is aligned and re-scaled to a size of $256 \times 256$ pixels.

\subsection{Authentication}
\label{sec:authentication}

Given a authentic $4$-second video clips for all unique identities, two reference sets are created with the VGG facial and Behavior-Net features. Define $F_i$ to be the $4096\times m_i$ real-valued matrix consisting of the VGG features for $m_i$ video clips of identity $i$. Similarly, define $B_i$ to be the $512 \times m_i$ real-valued matrix consisting of the Behavior-Net features for the same $m_i$ video clips, also of identity $i$. Each column of the matrices $F_i$ and $B_i$ contains the VGG and Behavior-Net features for a single video clip.

Given these reference sets, a previously unseen $4$-second video clip is authenticated as follows. First, extract the facial and Behavior-Net features, $\vec{f} \in \mathbb{R}^{4096}$ and $\vec{b} \in \mathbb{R}^{512}$. Next, find the identities, $i_f$ and $i_b$ in the reference sets with the most similar features using a cosine-similarity metric:
\begin{eqnarray}
    i_f ~=~ \argmax_i \{\max(\vec{f}^t \cdot F_{i})\} \qquad \mbox{and} \qquad
    i_b ~=~ \argmax_i \{\max(\vec{b}^t \cdot B_{i}) \}
\end{eqnarray}
With these matched identities, a video clip is classified as real or fake following two simple rules (see also Fig.~\ref{fig:overview}):
\begin{enumerate}
    \item A video clip is classified as real if the facial and Behavior-Net identities are the same, $i_f = i_b$, and if the facial similarity is above a specified threshold, $c_f >= \tau_f$, where $c_{f} = \max(\vec{f}^t \cdot F_{i_f})$ (i.e.,~a close facial match is found).
    \item A video clip is classified as fake if either
        \begin{enumerate}
            \item the matched identities are different, $i_f \neq i_b$, or 
            \item the facial similarity is below threshold, $c_f < \tau_f$.
        \end{enumerate}
\end{enumerate}
The rationale for the asymmetric treatment of the facial and Behavior-Net similarities is that in a face-swap deep fake, the facial identity of a person is modified but typically not the behavior. As a result, it is possible for a person's facial identity to be significantly different in a test video than in their reference videos, in which case, we should not be confident of the facial identity match.

\section{Results}
\label{sec:results}

We begin by describing the five datasets used for validation and analysis. We then describe the overall accuracy of detection followed by an analysis of robustness and relative importance of the appearance and behavioral features;

\subsection{Datasets}
\label{sec:datasets}

The world leaders dataset (WLDR) ~\cite{agarwal2019} consists of several hours of real videos of five U.S. political figures, their political impersonators, and face-swap deep fakes between each political figure and their corresponding impersonator. We augmented this dataset with five new U.S. political figures.

\begin{figure}[t]
    \begin{center}
        \begin{tabular}{ccc}
        Average & WLDR & FF  \\
        \includegraphics[width=0.3\textwidth]{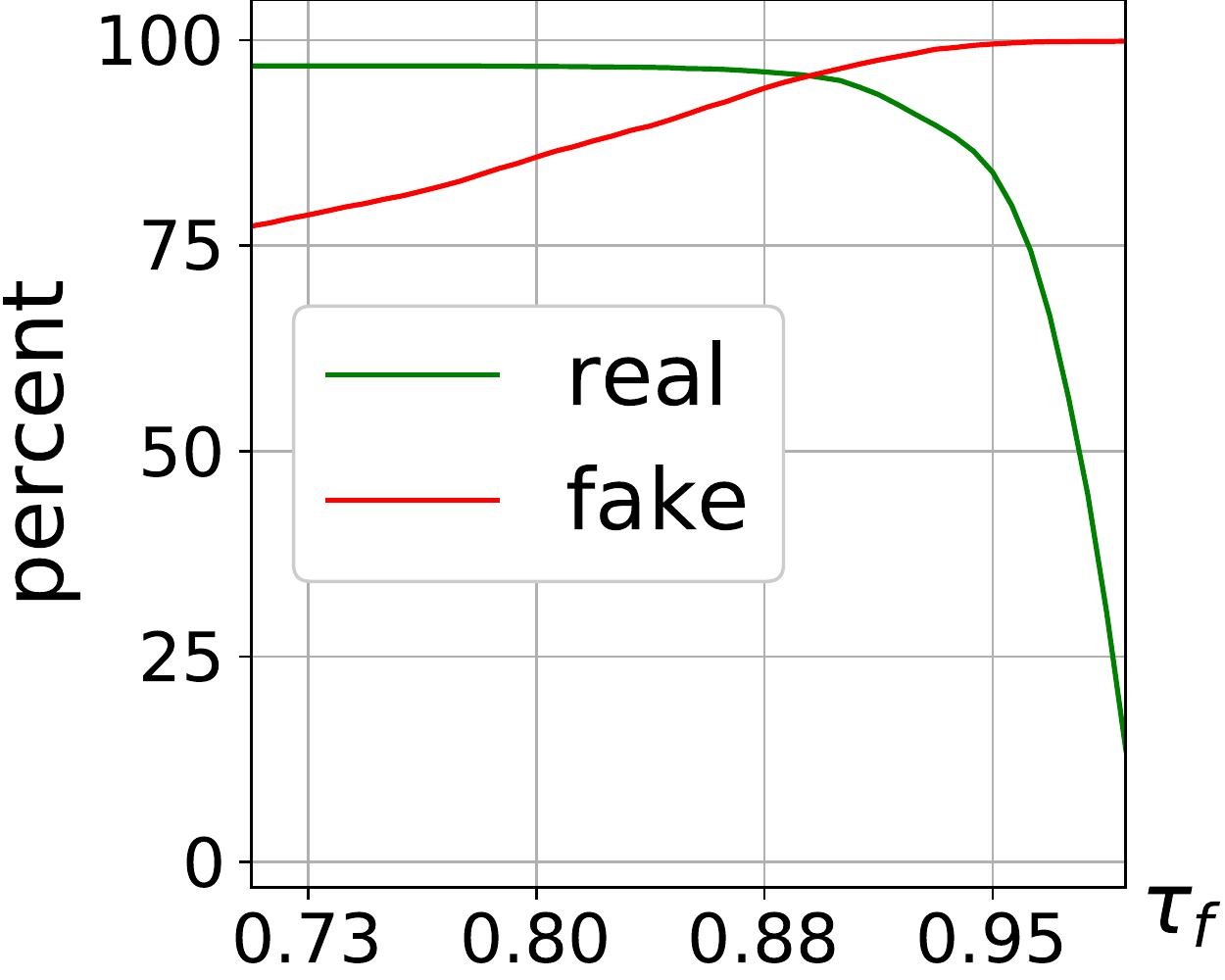} &        \includegraphics[width=0.3\textwidth]{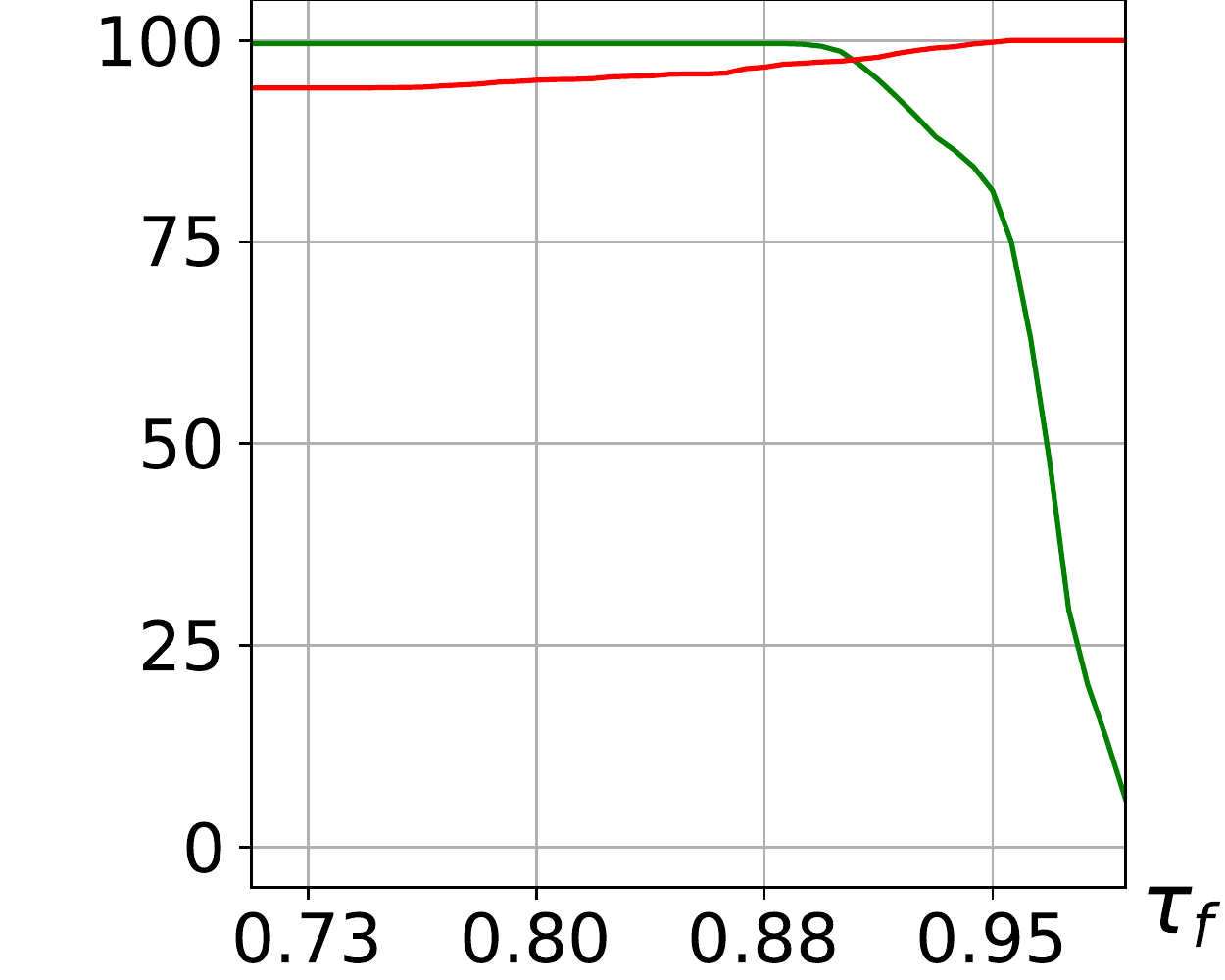} &
        \includegraphics[width=0.3\textwidth]{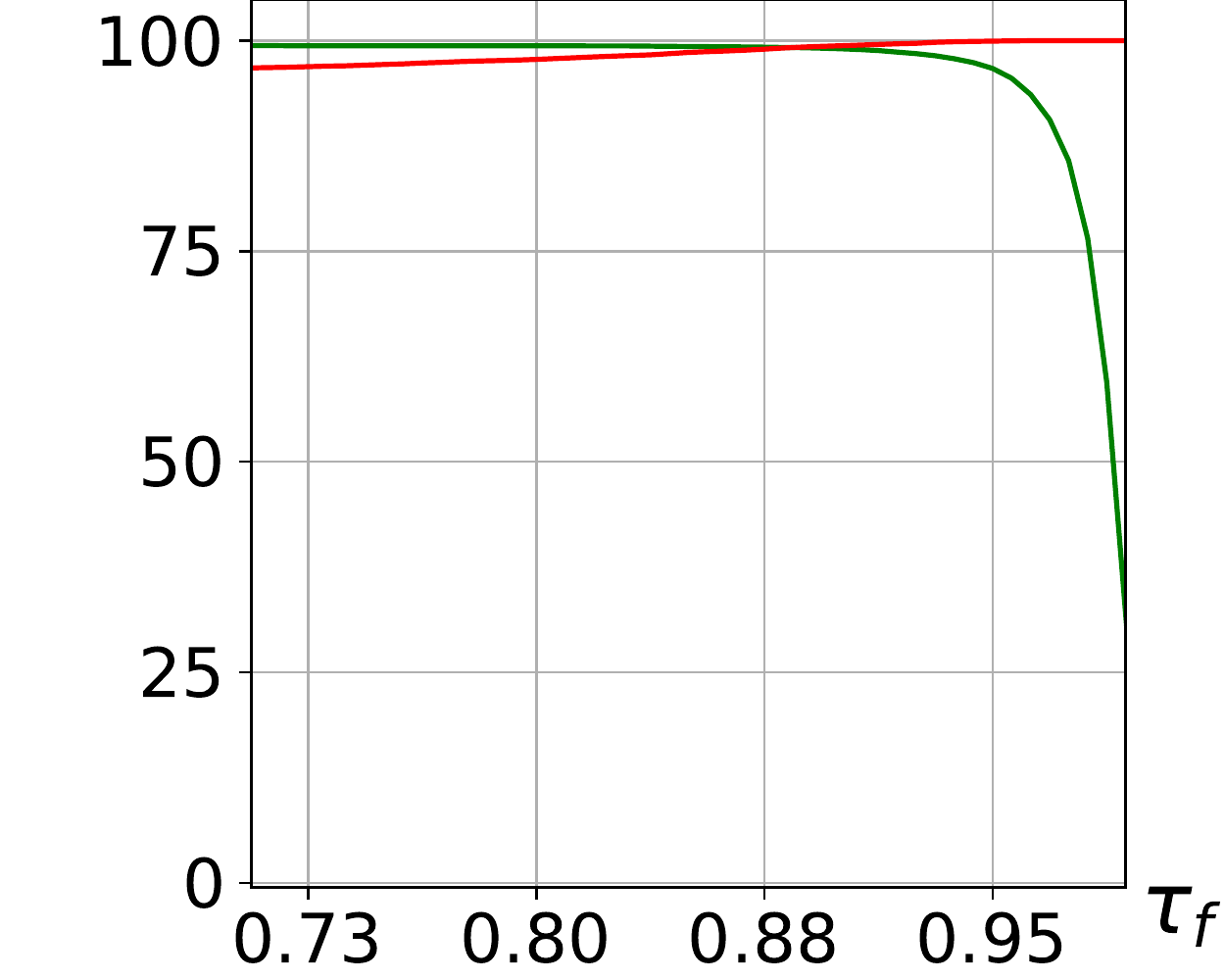}
        \\
        \\
        DFD & DFDC-P & CDF  \\
        \includegraphics[width=0.3\textwidth]{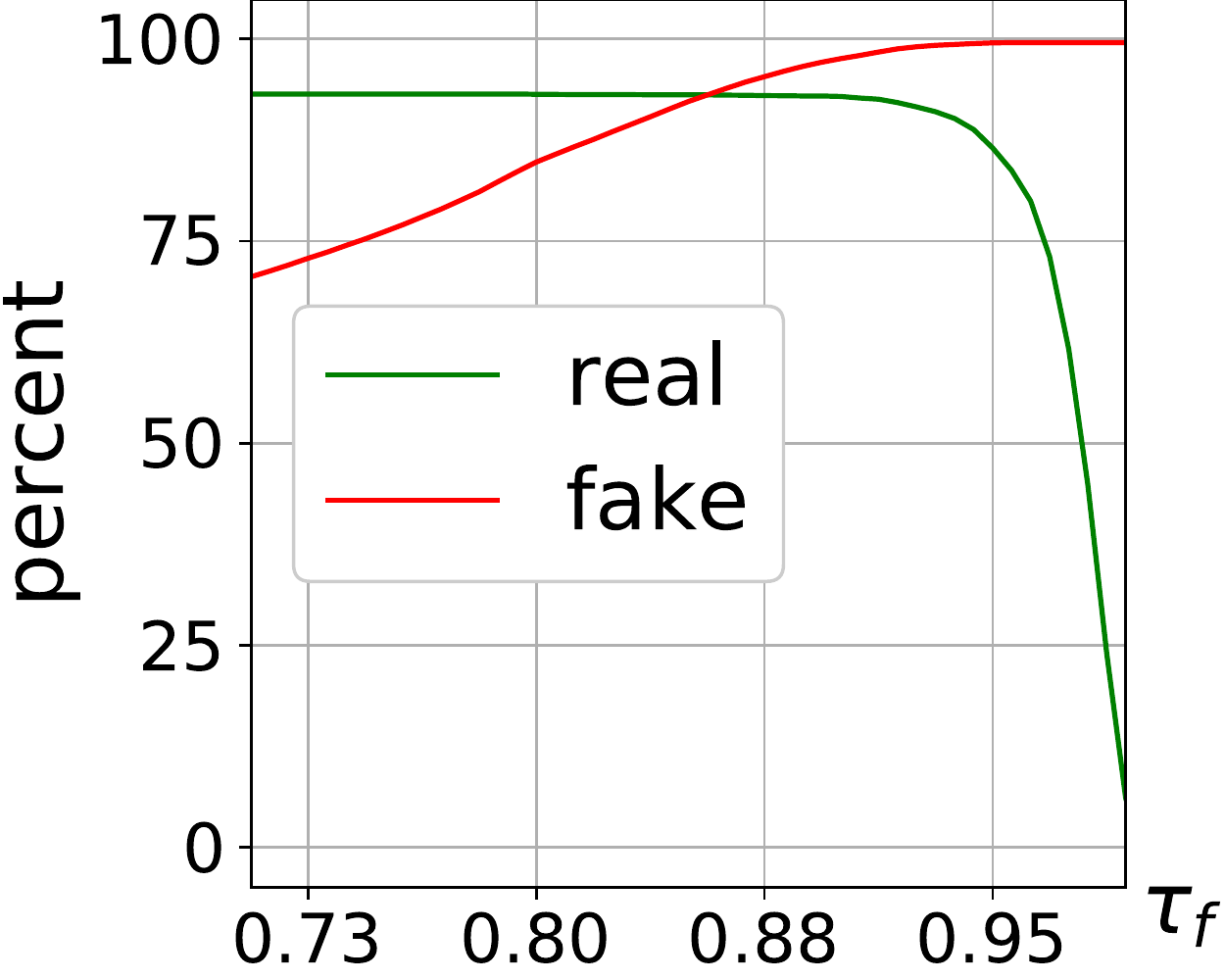} &
        \includegraphics[width=0.3\textwidth]{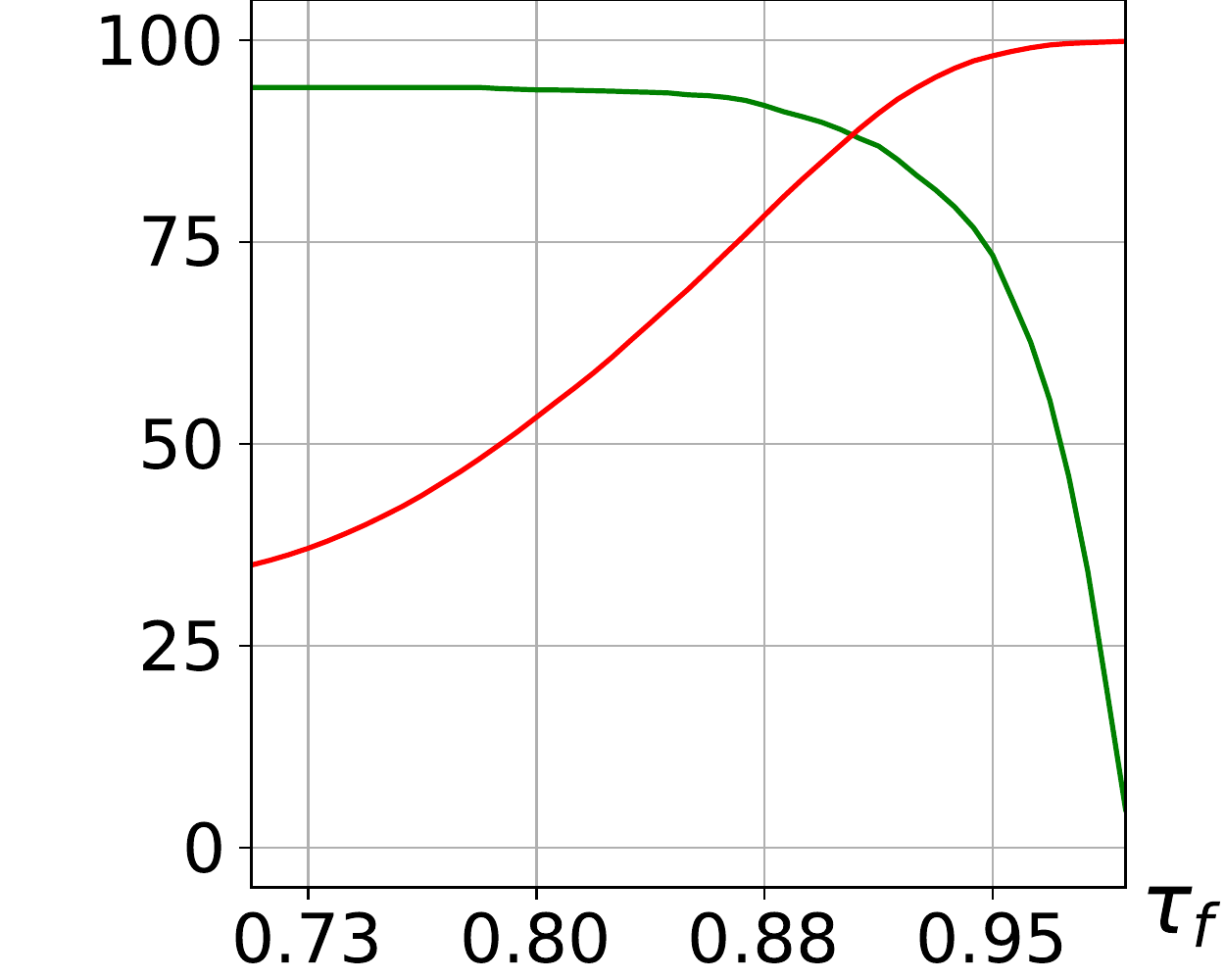} &
        \includegraphics[width=0.3\textwidth]{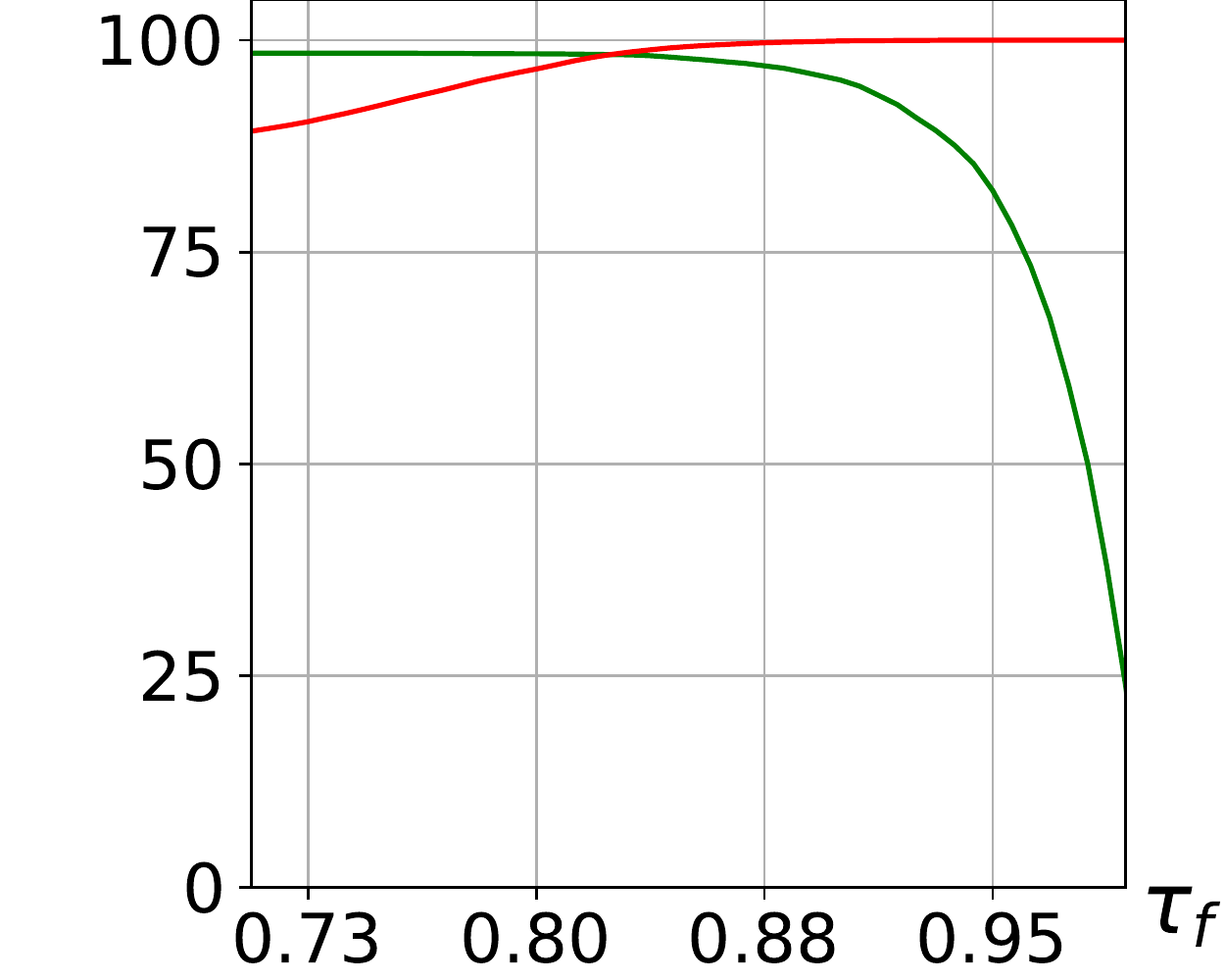} 
        \end{tabular} 
    \end{center}
    \vspace{-0.5cm}
    \caption{Shown are receiver operating curves (ROC) for each of five datasets and the average across all datasets (top-left panel). The green/red curves correspond to the accuracy of classifying real/fake videos. The horizontal axis corresponds to the VGG threshold ($\tau_f$).}
    \label{fig:ROC}
\end{figure}

The FaceForensics++ dataset (FF)~\cite{rossler2019} consists of $1000$ YouTube videos of $1000$ different people, mostly news anchors and video bloggers. Each video was used to create four types of deep fakes: DeepFake, FaceSwap, Face2Face, and Neural Textures. We only use the first two categories of fakes as only these are face-swap deep fakes. After removing videos with multiple people or with identities overlapping to other datasets, we were left with $990$ real videos and the corresponding $1980$ deep fake videos. 

The DeepFake Detection dataset (DFD)~\cite{dufour19} by Google/Jigsaw consists of $363$ real and $3068$ face-swap deep fakes of $28$ paid and consenting actors. Each individual was made to perform tasks like walking, hugging, talking, etc. in different expressions ranging from happy, to angry, neutral, or disgust. For our analysis, we selected only those videos where the individual was talking, resulting in $185$ real and $1577$ deep fake videos. 

The Deep Fake Detection Challenge Preview dataset (DFDC-P)~\cite{dolhansky2019} consists of $1131$ real and $4113$ face-swap deep fakes videos of $66$ consenting individuals of various genders, ages and ethnic groups. It is one of the largest deep fake dataset with videos of various quality, viewpoints, lighting conditions and scenes.  

The Celeb-DF (Ver. 2) dataset (CDF)~\cite{li2019celeb} is currently the largest publicly available deep-fake dataset. It is reported as containing $5639$ face-swap deep fakes generated from $590$ YouTube videos of $61$ celebrities speaking in different settings ranging from interviews, to TV-shows, and award functions (we, however, only identified $59$ unique identities in the downloaded dataset).

For each identity in the WLDR, DFD, and DFDC-P datsets, a random $80\%$ of the real videos are used for the reference set and the remaining $20\%$ are used for testing. In these three datasets there were sufficient videos of each individual in similar contexts. In contrast, the FF and CDF datasets had either only a small number of videos per individual or the context for each individual varied drastically. For these two datasets, therefore, we take a different approach to creating the reference/testing sets. In particular, each real video is divided in half, the first half of which is used for reference, and the second half used for testing. Similarly, we split each fake video in half, discard the first half and subject the second half to testing. The first half is discarded because the real counterpart of this video is used for reference, thus avoiding any overlap in utterances between the reference and testing. We recognize that this split is not ideal as video halves are not independent, but as we will see below, there is little difference in the results between the $80/20$ splits and these $50/50$ splits.

Each reference and testing video is re-saved at a frame-rate of $25$fps (and a ffmpeg quality of $20$). This consistent frame-rate allows us to partition each video into overlapping $4$-second clips, each of $100$ frames, with a $5$-frame sliding window.

\small
\begin{table}[t]
    \centering
    \begin{tabular}{l@{\hspace{1.2cm}}|@{\hspace{1.2cm}}c@{\hspace{1.2cm}}c@{\hspace{1.2cm}}c@{\hspace{1.2cm}}c@{\hspace{1.2cm}}c@{\hspace{1.2cm}}c}
                & Average & WLDR & FF & DFD & DFDC-P & CDF \\
        \hline
        real & $96.5\%$ & $99.6\%$ & $99.2\%$ & $93.1\%$ & $93.1\%$ & $97.6\%$ \\
        fake & $91.8\%$ & $95.8\%$ & $98.7\%$ & $93.2\%$ & $71.7\%$ & $99.4\%$ \\
        \hline 
        average & $94.2\%$ & $97.7\%$ & $98.9\%$ & $93.2\%$ & $82.4\%$ & $98.5\%$ \\
    \end{tabular}
    \vspace{0.2cm}
    \caption{Classification accuracies corresponding to the ROCs in Fig.~\ref{fig:ROC} at a fixed threshold of $\tau_f=0.86$. }
    \label{tab:accuracy}
\end{table}

\subsection{Identification}
\label{sec:identification}

Shown in Fig.~\ref{fig:ROC} are the receiver operating curves (ROC) for each of the five datasets enumerated in the previous section, along with the average across all datasets. The green/red curves correspond to the accuracy of classifying real/fake videos. The horizontal axis corresponds to the facial VGG threshold ($\tau_f$) used in determining if a video clip should be classified as real or fake (see Section~\ref{sec:authentication} and Fig.~\ref{fig:overview}).

As expected, as the threshold increases, the detection accuracy for fake (red) increases while the detection accuracy for real (green) decreases, particularly dramatically for threshold $\tau_f$ values that approach the maximum value of $1.0$. Recall that these accuracies are on a single $4$-second clip.

The cross-over accuracies in Fig.~\ref{fig:ROC} are $95.5\%$ (Average), $97.3\%$ (WLDR), $99.1\%$ (FF), $93.1\%$ (DFD), $88.4\%$ (DFDC-P), and $98.3\%$ (CDF). These cross-over points, however, come at varying $\tau_f$ threshold values. Shown in Table~\ref{tab:accuracy} is the detection accuracy, ranging from $82.4\%$ for DFDC-P to $98.9\%$ for FF, for a fixed threshold of $\tau_f=0.86$.

\begin{figure}[t]
    \centering
    \begin{tabular}{cccc}
     source & fake (face-swap) & target & \\
     \includegraphics[width=0.235\textwidth]{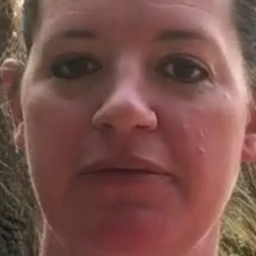} 
     & \includegraphics[width=0.235\textwidth]{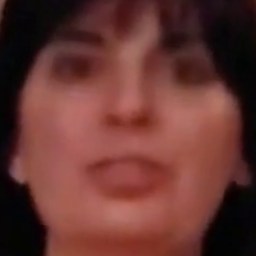} 
     & \includegraphics[width=0.235\textwidth]{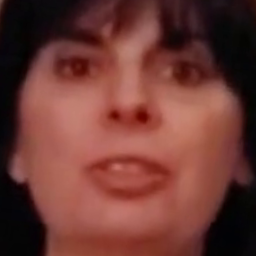}
     & \fbox{\includegraphics[width=0.235\textwidth]{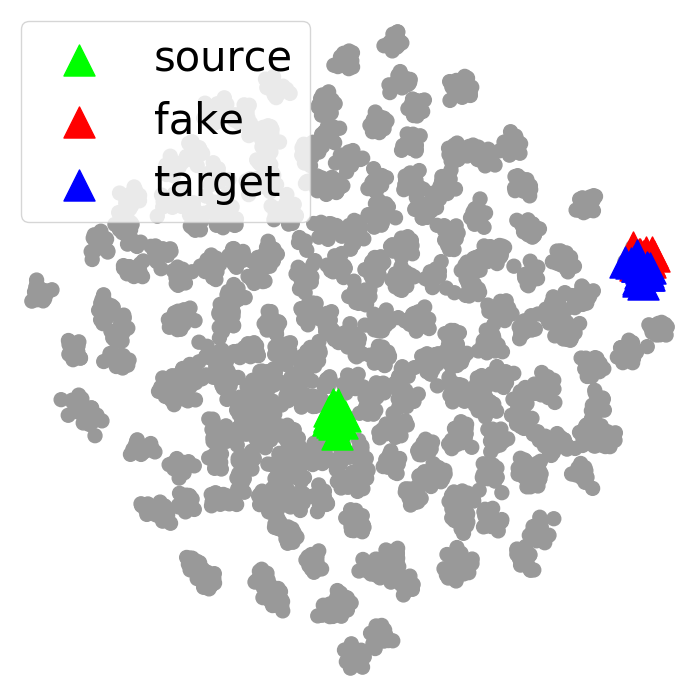}}
	\end{tabular}
    \caption{Shown is an example frame of a face-swap deep fake (second panel) from the DFDC-P dataset, in which the source identity (first panel) should be mapped onto the target (third panel), which is clearly not the case in this example. Shown in the fourth panel is the dimensionality-reduced visualization of the $4096$-D VGG features from all the identities (gray), source identity (green), target identity (blue), and the face-swap identity (red). This visualization shows that the source identity is not successfully mapped onto the deep fake (see also Fig.~\ref{fig:vgg_distribs}).}
    \label{fig:baddeepfake}
\end{figure}

\begin{figure}[t]
    \centering
    \begin{tabular}{cccccc}
    & WLDR & FF & DFDC-P & DFD & CDF \\
    \raisebox{0.6cm}{\rotatebox[origin=l]{90}{source}} &
    \includegraphics[width=0.18\textwidth]{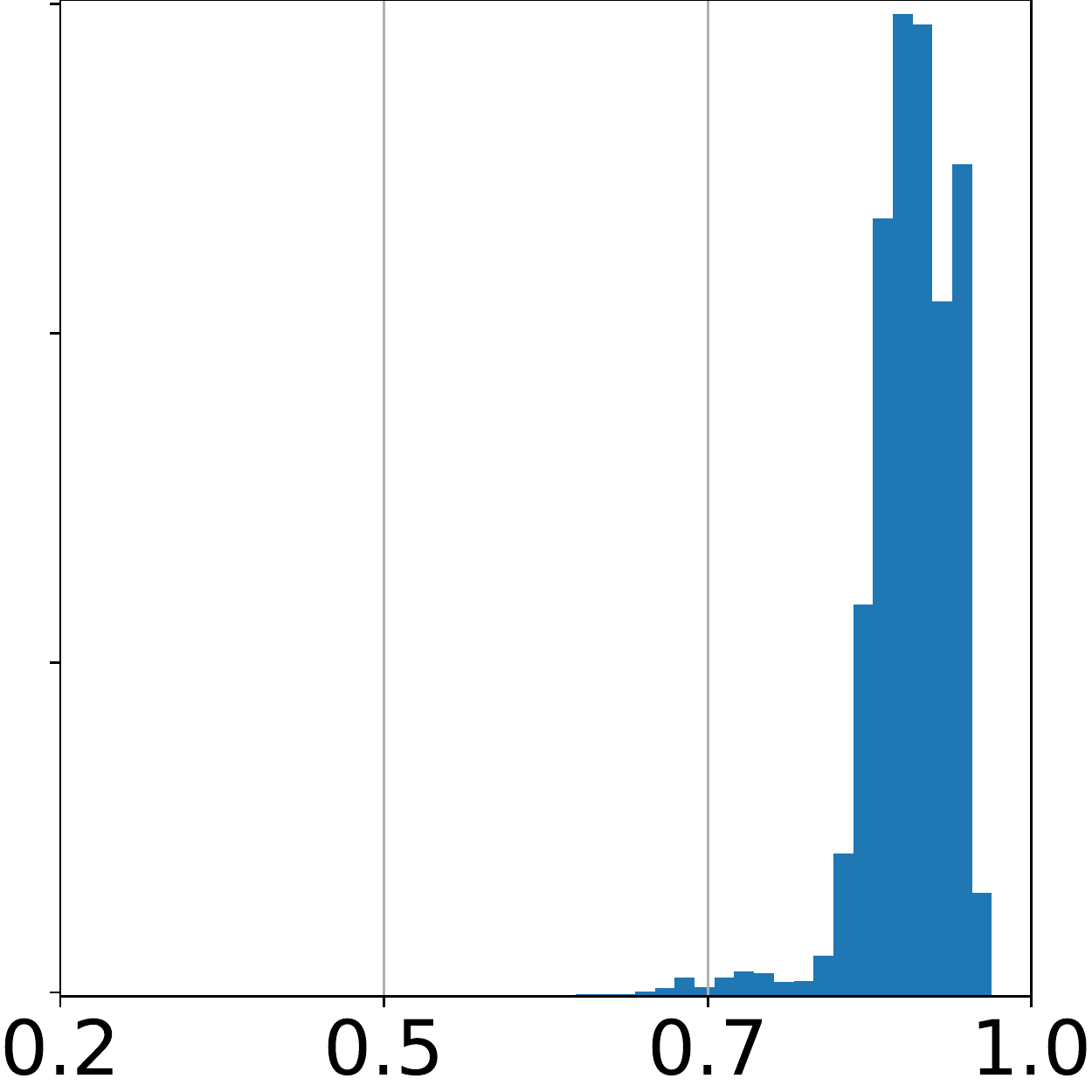}
    & \includegraphics[width=0.18\textwidth]{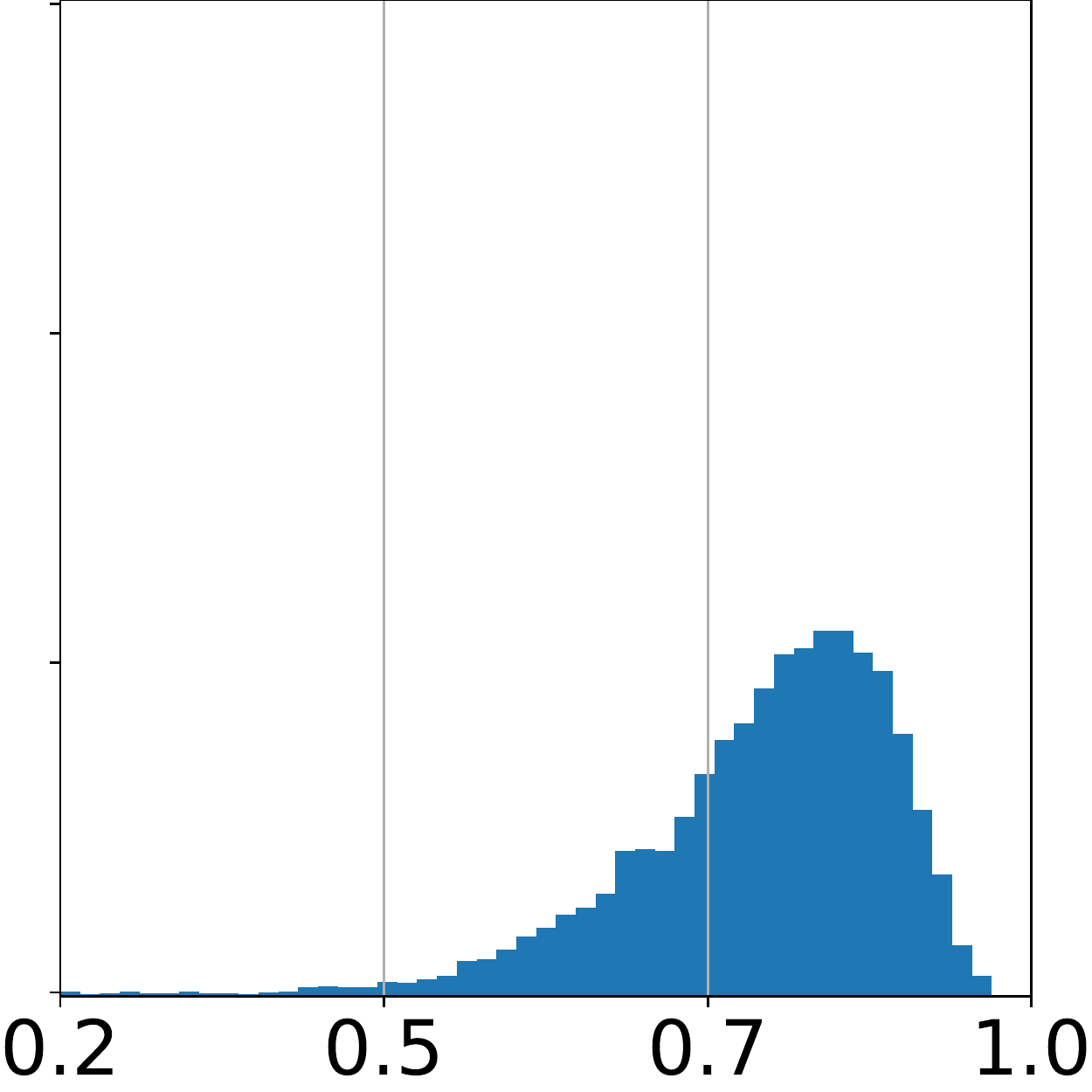}
    & \includegraphics[width=0.18\textwidth]{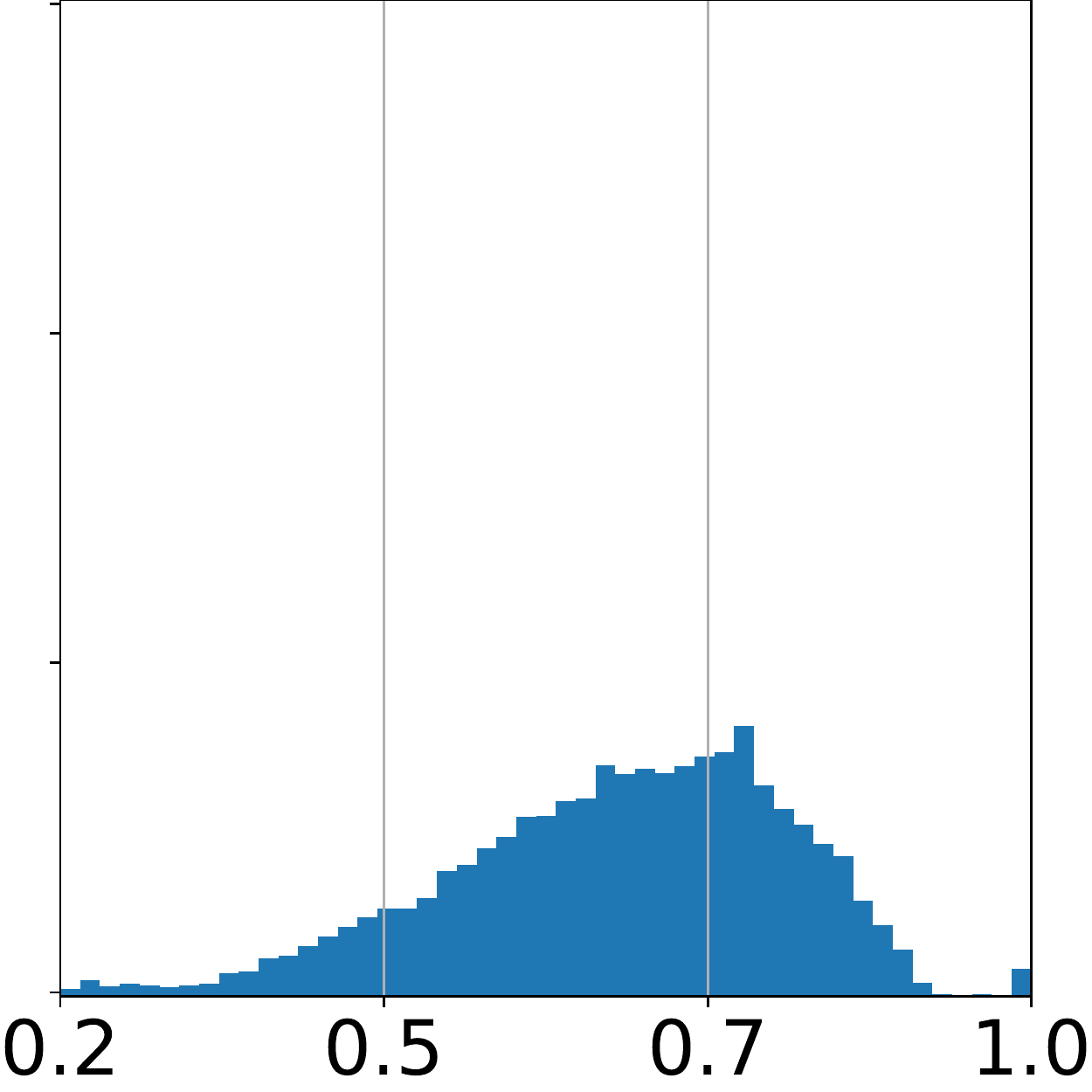} 
    & \includegraphics[width=0.18\textwidth]{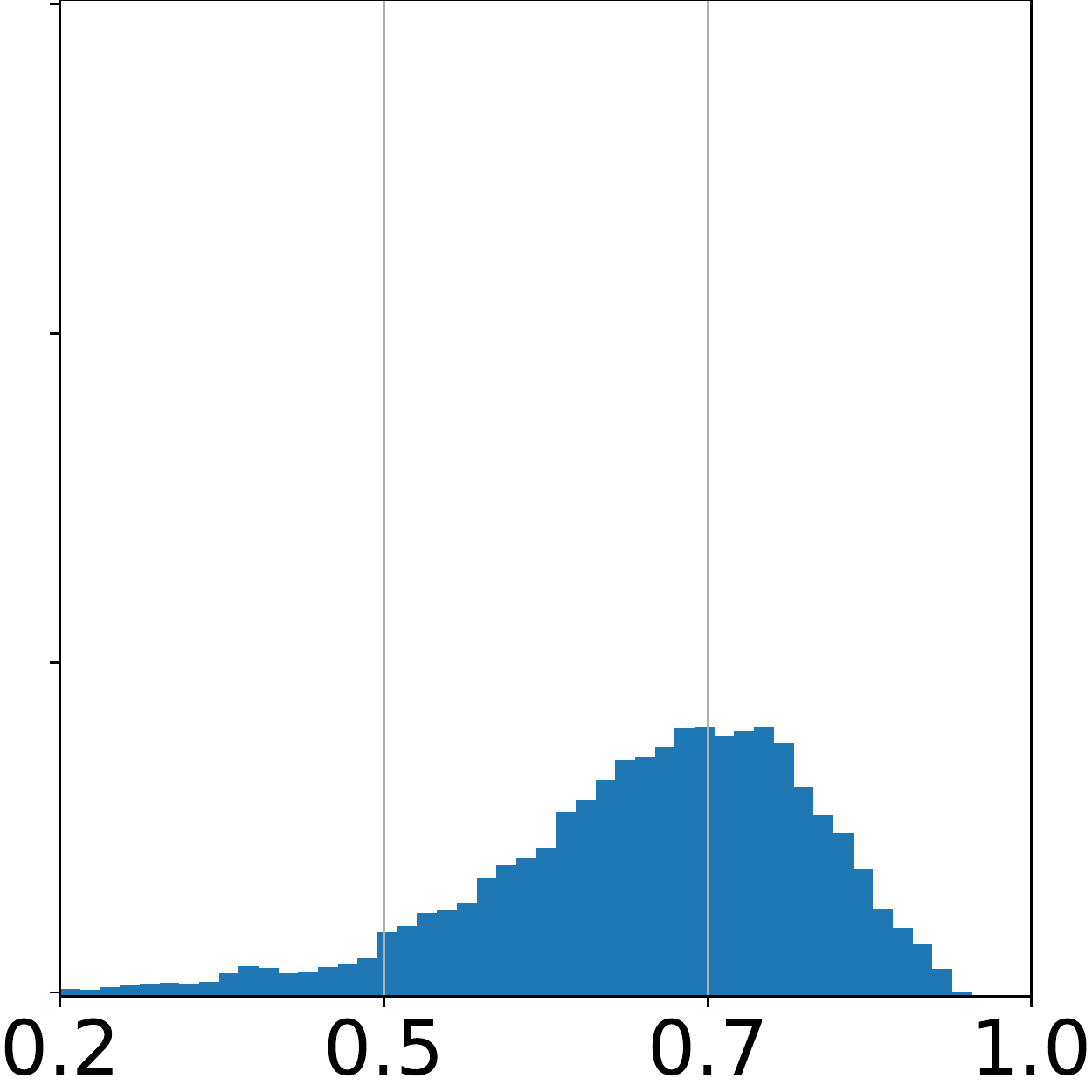}
    & \includegraphics[width=0.18\textwidth]{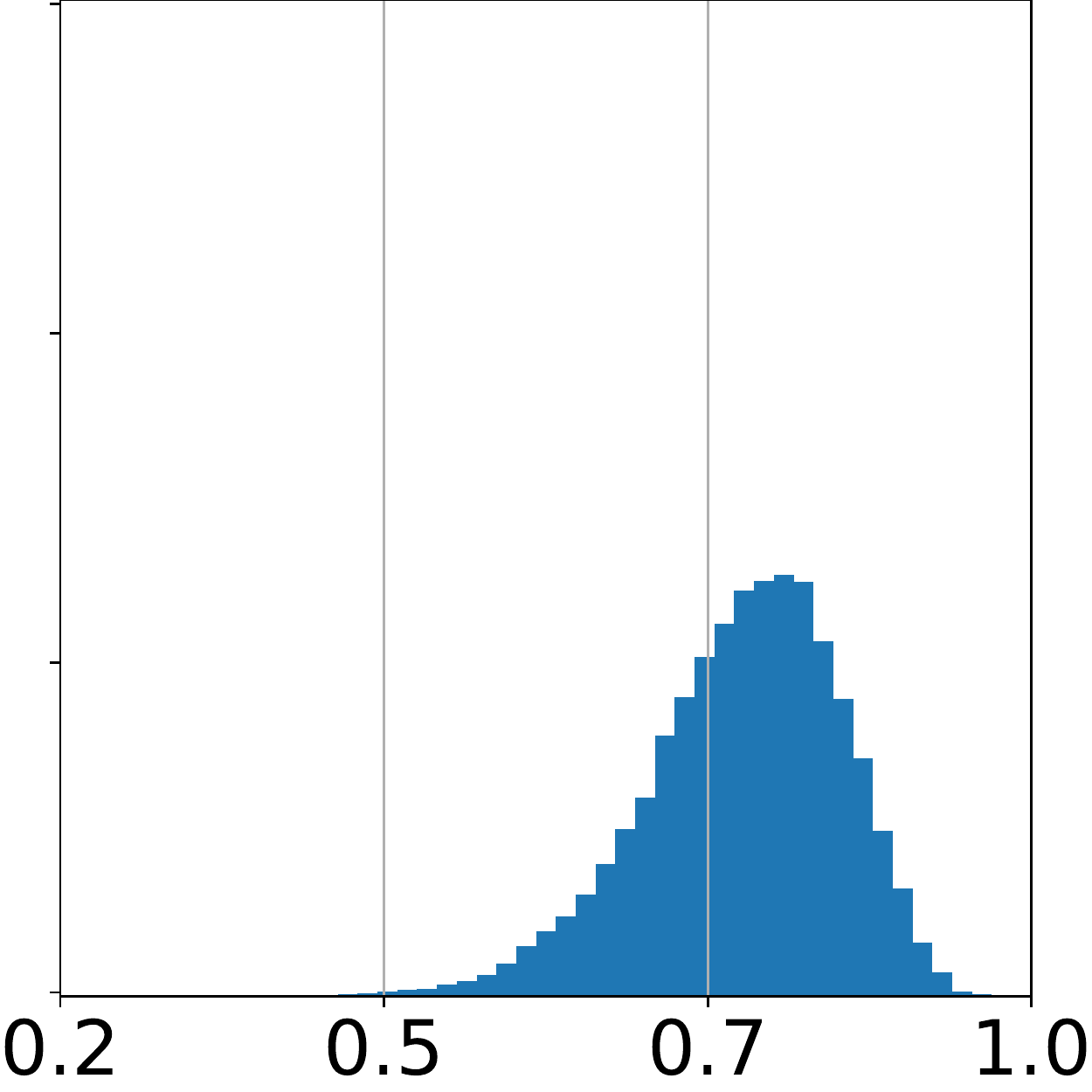} \\
    \raisebox{0.9cm}{\rotatebox[origin=l]{90}{target}} &
    \includegraphics[width=0.18\textwidth]{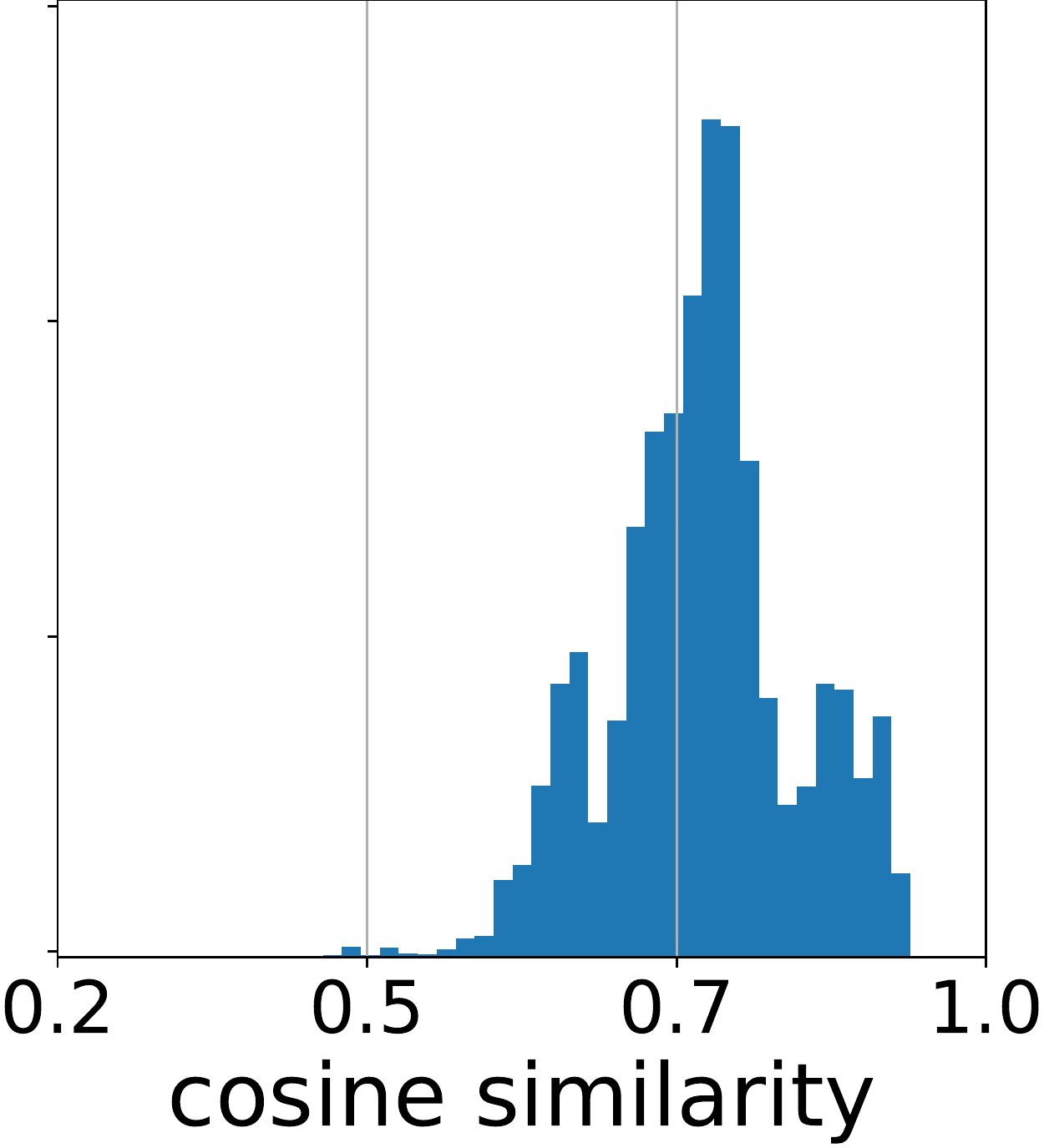} 
    & \includegraphics[width=0.18\textwidth]{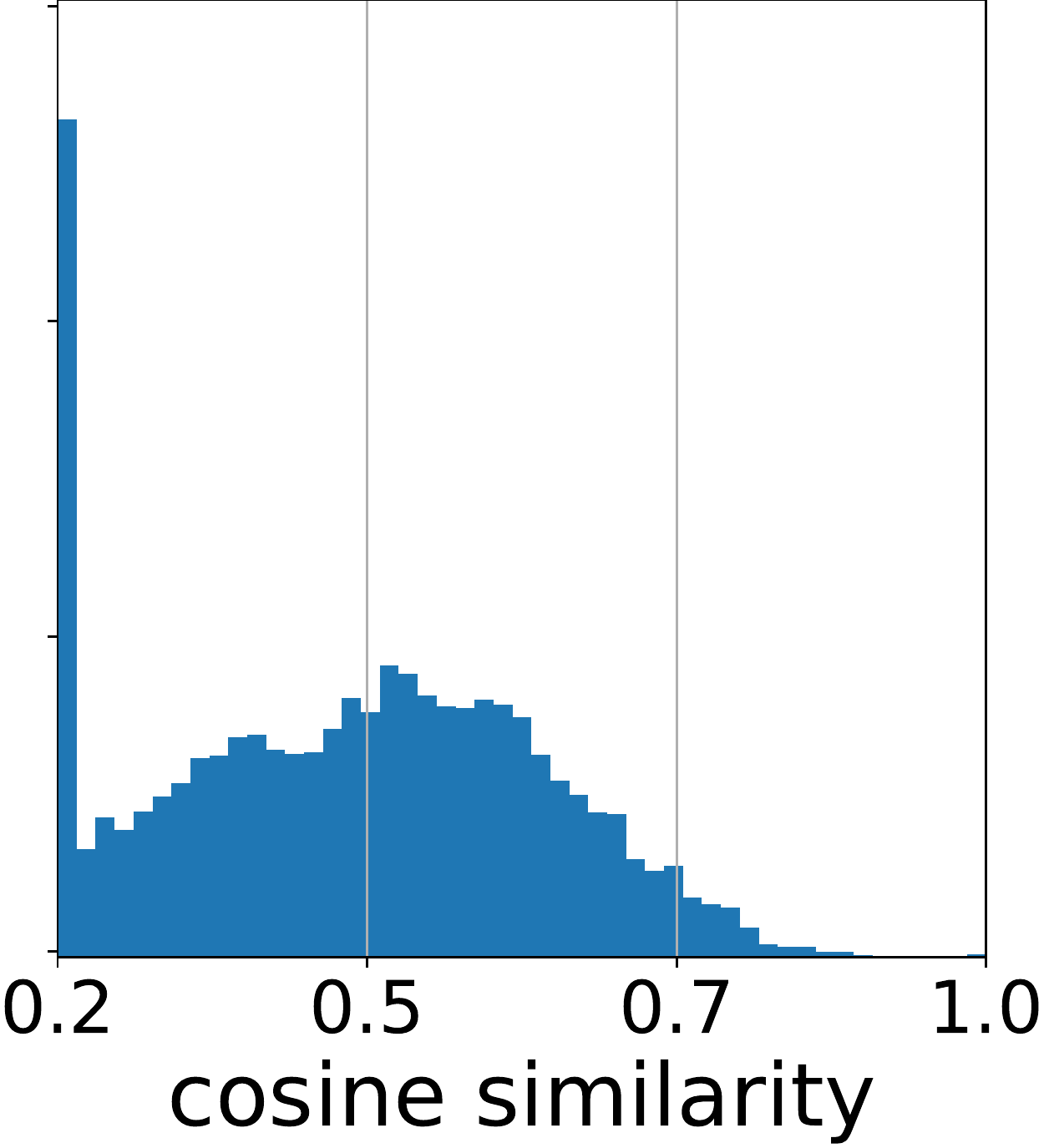}
    & \includegraphics[width=0.18\textwidth]{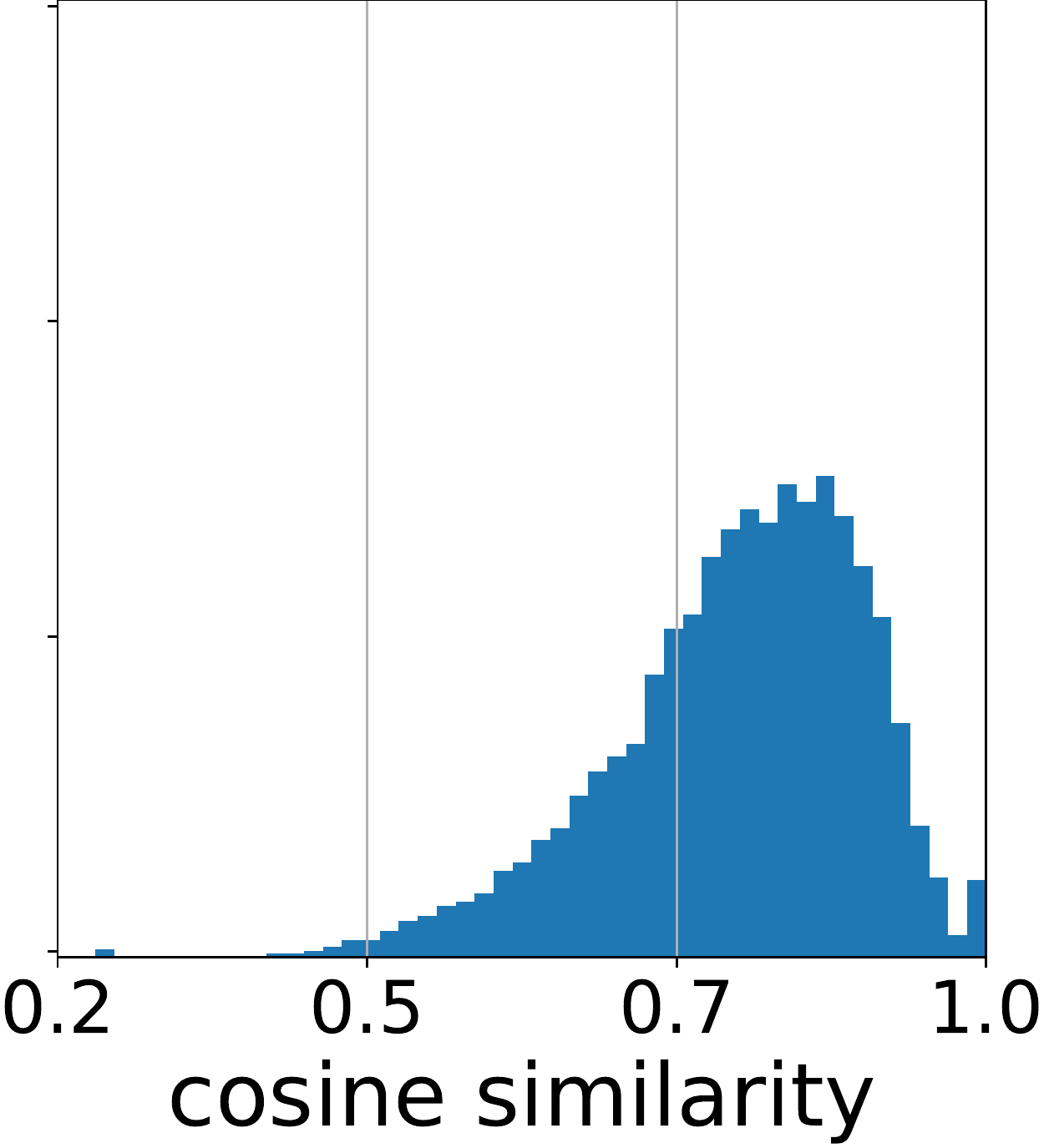}
    & \includegraphics[width=0.18\textwidth]{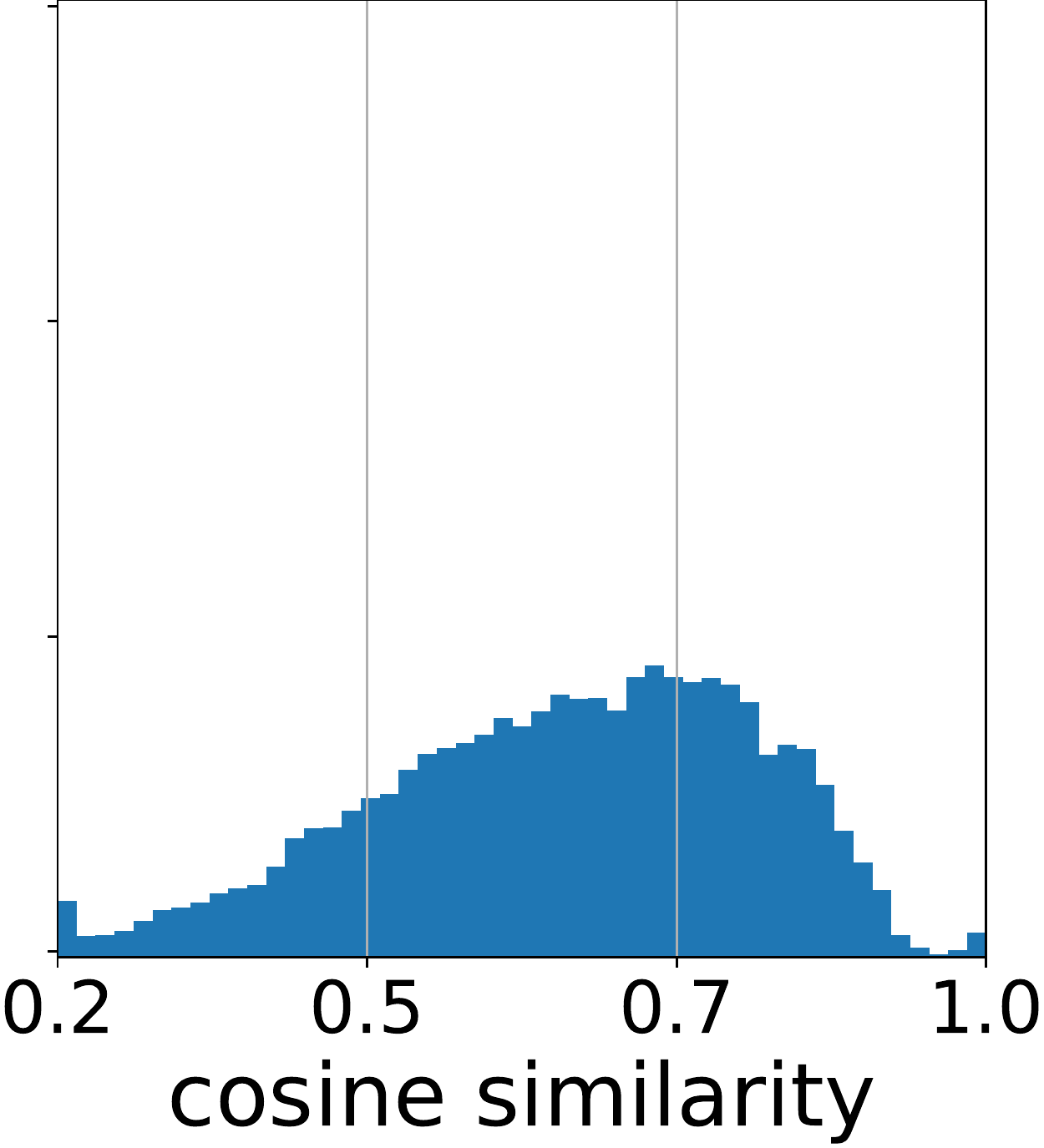}
    & \includegraphics[width=0.18\textwidth]{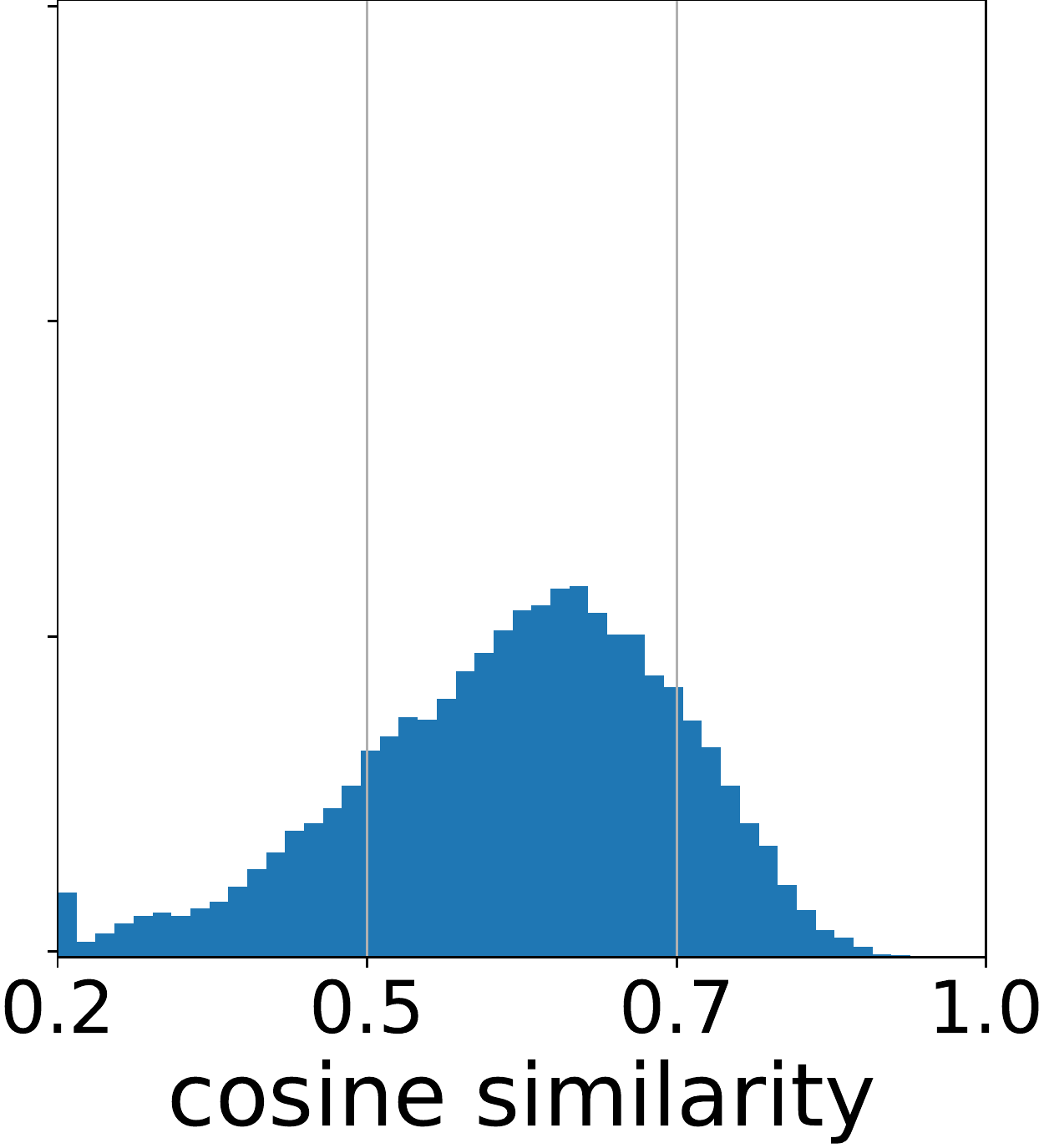}
    \end{tabular}
    \caption{The distributions in the first row correspond to the facial similarity between the faces in the source and fake videos (as computed by the cosine similarity between corresponding VGG features). The distributions in the second row correspond to the facial similarity between the faces in the target and fake videos. In a successful face-swap deep fake, the source to fake similarity will be higher than the target to fake similarity, as is the case for the WLDR dataset. For the DFDC-P dataset, however, these distributions are reversed (see also Fig.~\ref{fig:baddeepfake}).}
    \label{fig:vgg_distribs}
\end{figure}

Note that the accuracy for the DFDC-P is unusually low. This is because many of the fake videos in this dataset failed to convincingly map the facial appearance of the desired source identity into the target video. Shown in Fig.~\ref{fig:baddeepfake} is a representative example of this problem. Shown is one frame from the source video, one frame from the target video, and the corresponding frame from the face-swap deep fake video in which the source identity should be mapped into the target video. In this example drawn from the DFDC-P dataset, we can clearly see that the source identity was not mapped into the target video, but rather continues to look like the target. Shown in Fig.~\ref{fig:vgg_distribs} is confirmation that this problem persists throughout the DFDC-P dataset. In particular, shown in the first row are, for each dataset, the distribution of similarities in facial identities (as measured by the facial VGG cosine similarity) between all faces in the fake videos and their corresponding source identities. Shown in the second row is the similarity in facial identities all faces in the fake videos and their corresponding target identities. In a successful face swap, in which the identity in the target is replaced with that in the source, the facial similarity between the source and fake should be higher than the target and fake. Correspondingly, for each dataset, except DFDC-P, the average facial similarity of the fakes is higher relative to the source than the target. For the DFDC-P dataset, however, the fakes are on average closer to the target than the source. This difference accounts for the low accuracy on the DFDC-P dataset as both behavior and appearance of the fakes correspond to the target identity and are thus classified as real by our algorithm. Although this effect is most pronounced in the DFDC-P dataset, the DFD dataset also suffers from a similar problem, failing to convincingly map the source to the target identity. These failures justify our use of a confidence threshold in the facial similarity matching (case 2(b) in Section~\ref{sec:authentication}).

We next evaluate our detection algorithm against three in-the-wild, face-swap deep fake videos downloaded from YouTube. These three deep fakes were created using the following source and target combinations: 1) Steve Buscemi mapped onto Jennifer Lawrence~\footnote{\url{https://www.youtube.com/watch?v=VWrhRBb-1Ig}}; 2) Tom Cruise mapped onto Bill Hader~\footnote{\url{https://www.youtube.com/watch?v=r1jng79a5xc}}; and 3) Billie Eilish mapped onto Angela Martin~\footnote{\url{https://www.instagram.com/p/B6lXvJlIU92/}}. Because, only Jennifer Lawerence was already in our reference set (CDF), real videos for the other five identities were downloaded from YouTube to augment our reference set. This included three minutes of videos of Angela Martin from The Office and $20$ minutes of interview videos for each of Billie Eilish, Steve Buscemi, Bill Hader, and Tom Cruise. The accuracy rate for each of these face-swap deep fakes is $100\%$.

Lastly, shown in Table~\ref{tab:sota} is a comparison of our detection accuracy, measured using area under the curve (AUC), to six previous deep-fake detection schemes. Our scheme outperforms or is equal to previous approaches across all datasets. Note, however, that this is not a perfect comparison because our approach has access to a reference set of only real videos to compare against, as compared to these other fully-supervised approaches with access to real and fake reference videos.

\subsection{Analysis}
\label{sec:analysis}

Our Behavior-Net feature was designed to capture spatiotemporal behavior, while the VGG feature captures facial identity. Here we analyze our results in more detail to ensure that these two features are not entangled and that the Behavior-Net does in fact capture temporal properties not captured by the static FAb-Net features.

\begin{figure}[b]
    \begin{center}
        \begin{tabular}{c@{\hspace{0.4cm}}c@{\hspace{0.4cm}}c}
        (a) & (b) & (c) \\
        \includegraphics[width=0.3\textwidth]{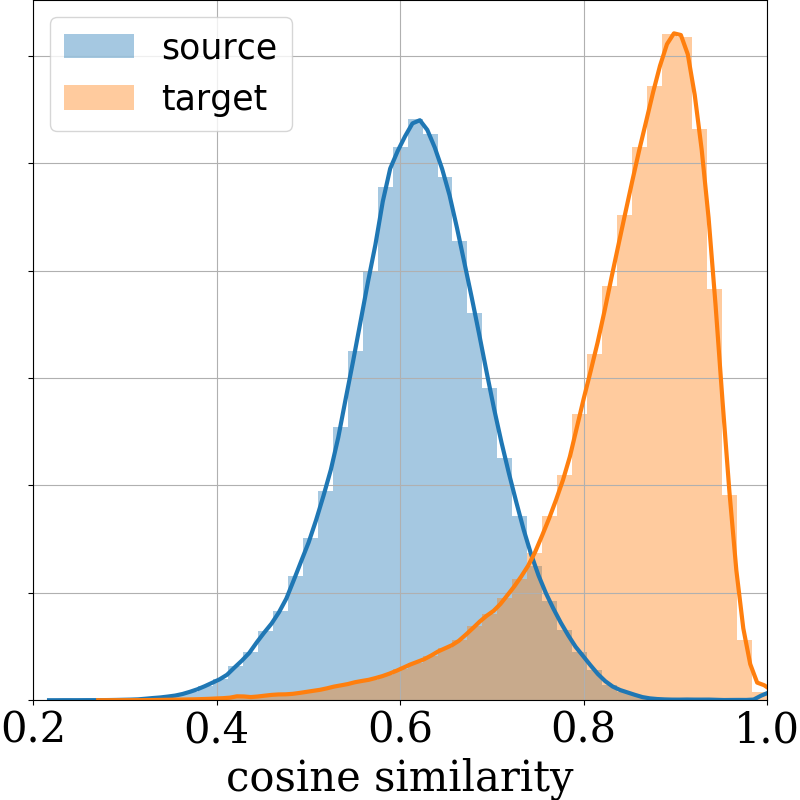}
        &\includegraphics[width=0.3\textwidth]{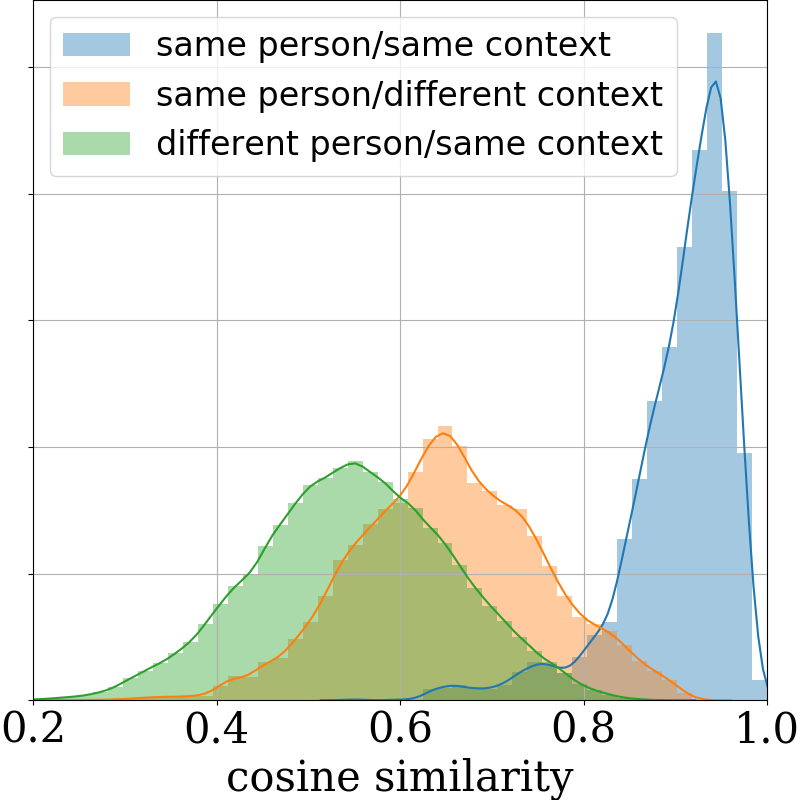} 
        &\includegraphics[width=0.3\textwidth]{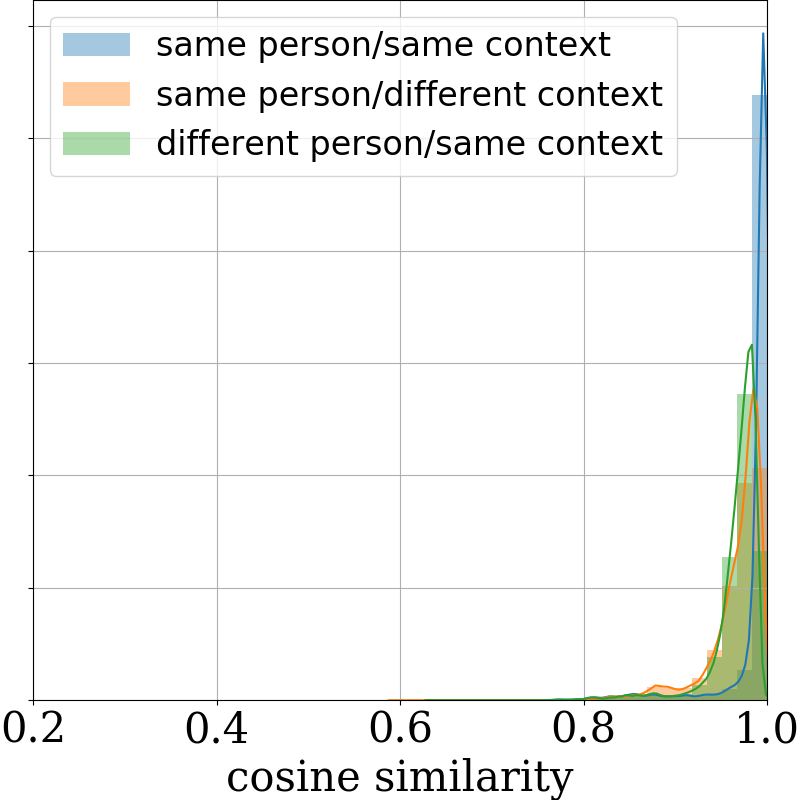}
        \end{tabular} 
    \end{center}
    \vspace{-0.5cm}
    \caption{Shown in panels (a) and (b) are the distributions of spatiotemporal behavior similarity, measured as the cosine similarity between Behavior-Net feature vectors. Shown in panel (c) is the distribution of spatial FAb-net similarity. See text for a detailed explanation of each panel.}
    \label{fig:entanglement}
\end{figure}

In the first analysis, we show that Behavior-Net does in fact capture behavior and not just a person's facial identity. Shown in Fig.~\ref{fig:entanglement}(a) are the distributions of Behavior-Net similarities between source (blue)/target (orange) identities relative to their face-swap deep fakes (recall that a face-swap deep fake is created by mapping an identity in a source video to a target video). The similarity of the target behavior relative to the face-swap deep fakes is much higher than the source, meaning that even though the facial identity in the deep fake matches the source, the behavioral identity still matches the target. This indicates that the Behavior-Net is capturing more information than just facial identity.

In the second analysis, we show that Behavior-Net captures identity-specific behaviors and not just identity-agnostic expressions or behaviors. This analysis is based on the real videos in the DFD dataset, where each of the $28$ actors were recorded talking in different contexts ranging from a casual conversation sitting on a couch to a speech at a podium. Each of these contexts captured a specific facial expression ranging from neutral, to angry, happy, and laughing. And, each of these contexts were recorded twice, once with a still camera and once with moving camera. Shown in Fig.~\ref{fig:entanglement}(b) are the distributions of Behavior-Net similarities between the same person in the same context (blue), the same person in different contexts (orange), and different people in the same context (green). When different people are recorded in the same context, we see that their Behavior-Net features are not similar, indicating that Behavior-Net captures identity-specific behaviors and not just specific contexts. At the same time, however, we see, that context can change an individual behavior (the orange vs. blue distributions). For example, a person is likely to have a different behavior when they are speaking casually to their friends as opposed to giving a formal speech to a large crowd. Nevertheless, our Behavior-Net captures identity-specific behaviors, albeit somewhat context dependent. Shown in Fig.~\ref{fig:entanglement}(c) are the same distributions as in panel (b) but for only the static FAb-Net features. The distributions for the same person in the same context (blue), the same person in different contexts (orange), and different people in the same context (green) are all nearly identical, revealing that the static FAb-Net features does not capture identity-specific information. 

In the third analysis, we analyse the amount of data required to build a reference set for an individual. For this analysis, the same reference set as before was used for the identities in FF, DFD, DFDC-P, and CDF. For the identities in the WLDR dataset (the only one with hours of video per person), the reference sets consists of between $1$ and $2000$ randomly selected $4$-second clips. With $2$, $30$, $50$, $100$, $1000$, and $2000$ video clips, the average detection accuracy for identities in the WLDR dataset are $65.4\%$, $92.2\%$, $93.2\%$, $94.0\%$, $97.3\%$, and $97.7\%$, respectively. This rapid increase in accuracy and leveling off shows that large reference sets are not needed, assuming, again, that the context in which the individual is depicted is similar.

In this fourth, and final, analysis, we analyse the robustness of classification against a simple compression laundering operation. The video clips in our reference and testing sets, Section~\ref{sec:identification}, are each encoded at a relatively high ffmpeg quality of qp=$20$ (the lower this value, the higher the quality). Each testing video clip was recompressed at a lower quality of qp$=40$ and classified against the original reference set. For the same threshold ($\tau_f=0.86$), the average detection accuracy remains high at $94.5\%$ (WLDR), $98.1\%$ (FF), $93.2\%$ (DFD), $80.9\%$ (DFDC-P), and $93.3\%$ (CDF). These results are almost identical to the high-quality videos in Table~\ref{tab:accuracy}.

\small
\begin{table}[t]
    \centering
    \begin{tabular}{l@{\hspace{1.cm}}|@{\hspace{1.cm}}c@{\hspace{1.cm}}c@{\hspace{1.cm}}c@{\hspace{1.cm}}c@{\hspace{1.cm}}c@{\hspace{1.cm}}c}
                & WLDR & FF & DFD & DFDC-P & CDF \\
        \hline
        Protecting World Leaders \cite{agarwal2019}  & 0.93 & -- & -- & -- & -- \\
        2-stream \cite{zhou2017} & -- & 0.70 & 0.52 & 0.61 & 0.53 \\
        XceptionNet-c23 \cite{li2019celeb} & -- & \textbf{0.99} & 0.85 & 0.72 & 0.65 \\
        Head Pose \cite{yang2019} & -- & 0.47  & 0.56 & 0.55 & 0.54 \\
        MesoNet \cite{afchar2018} & -- & 0.84 & 0.76 & 0.75 & 0.54 \\
        Face Warping \cite{li2018warping}  & -- & 0.80 & 0.74 & 0.72  &  0.56 \\ 
        \hline
        Ours: Appearance and Behavior & \textbf{0.99} & \textbf{0.99} & \textbf{0.93} & \textbf{0.95} & \textbf{0.99} \\
        \hline
    \end{tabular}
    \vspace{0.2cm}
    \caption{Comparison of our approach with previous work over multiple benchmarks~\cite{li2019celeb}. The reported values correspond to the AUC. Although not a perfect comparison due to significantly different underlying methodologies, our approach does perform well. The FF dataset in this comparison consists of the {\textit FaceSwap} and {\textit Deepfake} categories.}
    \label{tab:sota}
\end{table}
%
%

\section{Discussion}

We have developed a novel technique for detecting face-swap deep fakes. This technique leverages a fundamental flaw in these deep fakes in that the person depicted in the video is simply not the person that it purports to be. We have shown that a combination of a facial and behavioral biometric is highly effective at detecting these face-swap deep fakes. Unlike many other techniques, this approach is less vulnerable to counter attack and generalizes well to previously unseen deep fakes with previously unseen people.

Our forensic technique should generalize to so-called puppet-master deep fakes in which one person's facial expressions and head movements are mapped onto another person. These deep fakes suffer from the same basic problem as face-swap deep fakes in that the underlying behavior of the person is not that who it purports to be. As such, our combined facial and behavioral biometric should be able to detect these deep fakes.

We will, however, likely struggle to classify so-called lip-sync deep fakes in which only the mouth has been modified to be consistent with a new audio track. The facial identity and the vast majority of the behavior in these deep fakes will be consistent with the person depicted. To overcome this limitation, we seek to customize our behavioral model to learn explicit inconsistencies between the mouth and the rest of the face and/or underlying audio signal.

There is little question that the arms-race of synthesis and detection will continue. While it may not be possible to entirely stop the creation and distribution of deep fakes, our, and related approaches, promise to make the creation of convincing deep fakes more difficult and time consuming. This will eventually take it out of the hands of the average person and relegate it to the hands of a fewer and fewer experts. While the threat of deep fakes will remain, this will surely be a more manageable threat.

\section{Acknowledgement}
The PI's research group (Farid) is partially supported with funding from the Defense Advanced Research Projects Agency (DARPA FA8750-16-C-0166). The views, opinions, and findings expressed are those of the authors and should not be interpreted as representing the official views or policies of the Department of Defense or the U.S. Government. The PI's research group is also partially supported by Facebook. There is no collaboration between Facebook and DARPA. We thank Yipin Zhou for her help in data collection.

\clearpage
%
\bibliographystyle{splncs04}
\bibliography{main}
\end{document}